\pdfoutput=1

\documentclass[11pt]{article}

\usepackage[final]{acl}

\usepackage{times}
\usepackage{latexsym}
\usepackage{CJKutf8}
\usepackage[T1]{fontenc}

\usepackage[utf8]{inputenc}

\usepackage{microtype}

\usepackage{inconsolata}
\usepackage{graphicx}
\usepackage{tcolorbox}
\usepackage{enumitem}
\usepackage{amsmath}
\usepackage{hyperref}
\usepackage{url}
\usepackage{booktabs}
\usepackage{amsfonts}
\usepackage{tabularx} 
\usepackage{ragged2e} 
\usepackage{setspace}
\usepackage{xcolor}
\usepackage{subfigure}
\usepackage{multirow}
\usepackage{color, colortbl}
\definecolor{mygray}{gray}{0.9}
\usepackage[normalem]{ulem}
\useunder{\uline}{\ul}{}
\title{Translation with Thought: Difficulty-Adaptive Reasoning via Reinforcement Learning for Multi-Domain Machine Translation}

\author{
Yongshi Ye\textsuperscript{1,3},
Biao Fu\textsuperscript{2,3,}\thanks{\,\,Corresponding authors.},
Chongxuan Huang\textsuperscript{2,3},
Yidong Chen\textsuperscript{2,3},
Xiaodong Shi\textsuperscript{1,2,3,}\footnotemark[1]
\\[0.5em]
\textsuperscript{1}Institute of Artificial Intelligence, Xiamen University \\
\textsuperscript{2}School of Informatics, Xiamen University \\
\textsuperscript{3}Key Laboratory of Digital Protection and Intelligent Processing of Intangible Cultural \\ Heritage of Fujian and Taiwan (Xiamen University), Ministry of Culture and Tourism\\
\texttt{\{yeyongshi,biaofu\}@stu.xmu.edu.cn,mandel@xmu.edu.cn}
}

\begin{document}
\maketitle

\begin{abstract}
Multi-domain machine translation (MDMT) poses a unique challenge due to varying levels of linguistic complexity across domains. Inspired by human translators' ability to adapt reasoning effort based on difficulty, we propose \textbf{\texttt{TwT}} (\textbf{T}ranslation \textbf{w}ith \textbf{T}hought), a resource-rational framework that learns to modulate inference between intuitive and deliberate reasoning.
\textbf{\texttt{TwT}} is trained in two stages: (1) supervised fine-tuning on difficulty-aware long chain-of-thought traces distilled from DeepSeek-R1 and rewritten by GPT-4o to reflect human-like reasoning economy, and (2) reinforcement learning with a hybrid reward to optimize translation quality and reasoning efficiency.
Evaluated on 15 benchmarks spanning in-domain and out-of-domain settings, as well as 3 seen and 59 unseen languages, with ablations across three backbone models, \textbf{\texttt{TwT-7B}} and \textbf{\texttt{TwT-14B}} outperform much larger SOTA reasoning models in translation quality, while reducing token usage by 32–60\%. These results confirm that aligning translation behavior with cognitive principles enables robust generalization, high translation quality, and efficient reasoning in MDMT.
\end{abstract}

\begin{figure*}[th]
    \centering
    \includegraphics[trim=10 0 10 300,clip,angle=0.2,width=0.9\textwidth]{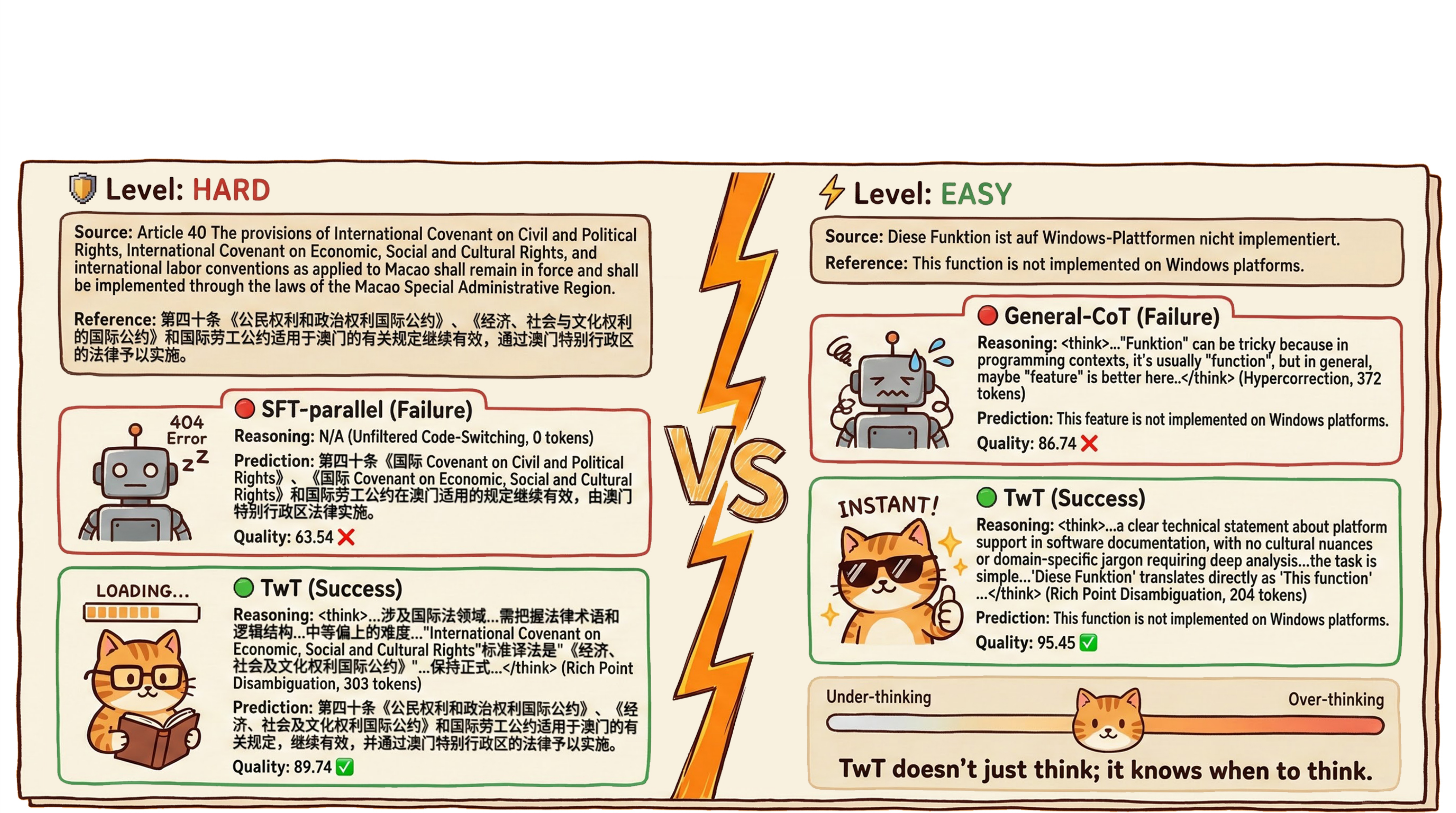}
    \caption{Case Study of Adaptive Thought. \textbf{\texttt{TwT}} switches between System 1 and System 2 based on complexity.}
    \label{fig:sys1_sys2_case}
\end{figure*}

\section{Introduction}
\label{sec:intro}

Multi-domain machine translation (MDMT) remains a core challenge for language models due to significant variation in terminology, syntax, and style across domains. A key difficulty lies in the uneven distribution of complexity: some inputs are routine, while others require deeper reasoning to resolve ambiguity or domain-specific constructs. 
However, most MT systems translate all inputs uniformly, lacking mechanisms to adjust inference effort based on domain-specific complexity~\citep{li2025testtimescalingreasoningmodels,liu2025newtrendsmodernmachine}. 
This contrasts with human translators, who adapt reasoning effort to input difficulty~\citep{hvelplund2011allocation,gile2020translation}. They typically rely on fast, intuitive processing (System~1) for familiar content and slower, deliberate reasoning (System~2) when encountering Rich Points~\citep{agar1994language}, such as ambiguous terminology, complex syntax, or cultural disparities. Because the density of such Rich Points varies across domains and correlates with translation difficulty~\citep{lacruz2017cognitive}, current MT systems still largely lack this adaptive reasoning ability.

From the perspective of reasoning allocation, existing MT approaches fall into two extremes. On one end, standard large language model (LLM)-based translators operate purely in System~1 mode: trained via supervised fine-tuning (SFT) on large-scale parallel corpora~\citep{xu2024a}, they produce fluent outputs without explicit reasoning. While efficient, these models struggle with Rich Points and degrade in out-of-domain (OOD) or low-resource settings.
Recent efforts have introduced Chain-of-Thought (CoT) prompting into translation~\citep{wang2025drtdeepreasoningtranslation}, but this does not fundamentally solve the problem, because the same reasoning pattern is applied uniformly regardless of input difficulty.
Conversely, the emergence of large reasoning models (LRMs), such as DeepSeek-R1~\citep{deepseekai2025deepseekr1}, represents a shift to the opposite extreme—an overcommitment to System~2. Reinforcement learning (RL) is often used to train these models to generate Long CoT traces, applying reasoning uniformly across inputs. This raises a natural question: \textit{Can RL serve as a bridge to align the model's reasoning trajectory with the human translation process?}

To investigate this, we conduct two preliminary experiments (Section~\ref{sec:preliminary}). 
The first examines \textit{Pure RL}, where RL is applied directly to a base model without SFT on annotated reasoning traces. We find that the model rapidly collapses into repetitive, shallow templates, failing to develop domain-specific reasoning behaviors. 
The second explores \textit{RL with SFT}, which fine-tunes on CoT traces before RL. While this setup produces longer reasoning, it lacks control over when such reasoning is needed, leading to verbose traces even for simple inputs. This indiscriminate reasoning may help reveal Rich Points, but often results in overthinking and excessive token usage, reducing efficiency and human alignment.
Despite this, most reasoning-based MT methods still adopt either \textit{Pure RL}~\citep{feng2025mtr1zeroadvancingllmbasedmachine} or \textit{RL with SFT}~\citep{wang2025extransmultilingualdeepreasoning}, differing mainly in reward design. However, few attempt to align reasoning effort explicitly with input difficulty.

Inspired by human cognitive flexibility, we propose \textbf{\texttt{TwT}} (\textbf{T}ranslation \textbf{w}ith \textbf{T}hought), a resource-rational framework for MDMT that learns to allocate inference effort based on input difficulty. 
As shown in Figure~\ref{fig:sys1_sys2_case}, \textbf{\texttt{TwT}} dynamically shifts its reasoning behavior according to input difficulty, using concise reasoning for routine inputs and deeper reasoning for domain-specific challenges.
To implement this, \textbf{\texttt{TwT}} follows the RL with SFT pipeline.
In the cold-start stage, it performs multi-agent distillation to construct difficulty-adaptive reasoning traces across domains: a domain-specialized teacher (DeepSeek-R1) generates diverse reasoning traces, and GPT-4o assesses input difficulty via Rich Points, rewriting the traces to match the appropriate reasoning depth.
This process equips our student model with domain-sensitive reasoning and human-like inference modulation, supporting resource-rational translation.
In the RL stage, we optimize the adaptive reasoning behavior seeded during cold-start by rewarding high-quality translations. Our hybrid reward combines translation quality metrics (BLEU and COMET) with a repetition penalty, 
guiding the model toward 
efficient, domain-adaptive reasoning through outcome-driven learning.

We conduct a comprehensive evaluation of \textbf{\texttt{TwT}} on 15 benchmarks across in-domain and OOD settings, as well as 3 seen and 59 unseen languages, and perform ablation studies on three different backbone models to assess generalization across both domain and linguistic axes. 
Our results demonstrate that \textbf{\texttt{TwT}} achieves performance competitive with or superior to SOTA LRMs (e.g., DeepSeek-R1, OpenAI-o1) and surpass strong MT-specialized baselines, while reducing token usage by 32--60\%. 
Empirical analysis confirms that \textbf{\texttt{TwT}} effectively modulates reasoning effort according to task difficulty, leading to more coherent reasoning processes and more accurate translations.
This validates the core intuition behind \textbf{\texttt{TwT}}: aligning reasoning effort with input difficulty yields both efficiency and quality gains.

\section{Related Work}
Recent MT studies increasingly explore explicit reasoning to improve translation quality, starting with shallow strategies such as disambiguation, domain recognition, and self-reflection~\cite{chen-etal-2024-dual,feng-etal-2025-tear,wang-etal-2024-taste,hu-etal-2024-large-language}.
To support deeper reasoning, recent studies collect Long CoT traces via MCTS~\cite{zhao2024marcoo1openreasoningmodels} or multi-agent workflows~\cite{wang2025drtdeepreasoningtranslation}, then apply SFT. These traces emulate human translation workflows, improving both performance and interpretability~\cite{chen2025evaluatingo1likellmsunlocking,liu2025newtrendsmodernmachine}.
More recently, RL has emerged as a reasoning enhancer~\cite{deepseekai2025deepseekr1}. Several approaches optimize translation reasoning with verifiable rewards: R1-T1 uses COMET-based signals~\cite{he2025r1t1fullyincentivizingtranslation}, MT-R1-Zero combines rule-based and neural metrics~\cite{feng2025mtr1zeroadvancingllmbasedmachine}, DeepTrans employs external LLMs~\cite{wang2025deepreasoningtranslationreinforcement}, and ExTrans adds exemplar-based guidance~\cite{wang2025extransmultilingualdeepreasoning}.
However, these methods overlook cognitive alignment; we model human-like reasoning to improve MDMT efficiency and quality.

\section{Preliminary Analysis}
\label{sec:preliminary}
\subsection{Reasoning Collapse}
\label{sec:pure_rl_challenge}

\begin{figure*}[t]
\centering
\subfigure[Pure RL training with KL regularization.]{
\label{fig:zero_w_kl}
\includegraphics[width=0.87\textwidth]{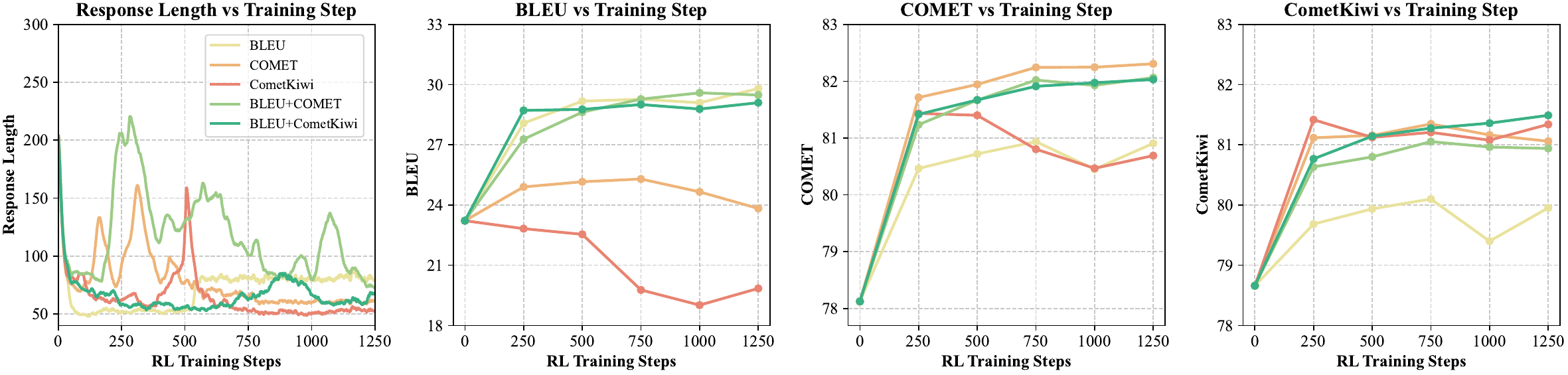}
}
~
\subfigure[Pure RL training without KL regularization.]{
\label{fig:zero_wo_kl}
\includegraphics[width=0.87\textwidth]{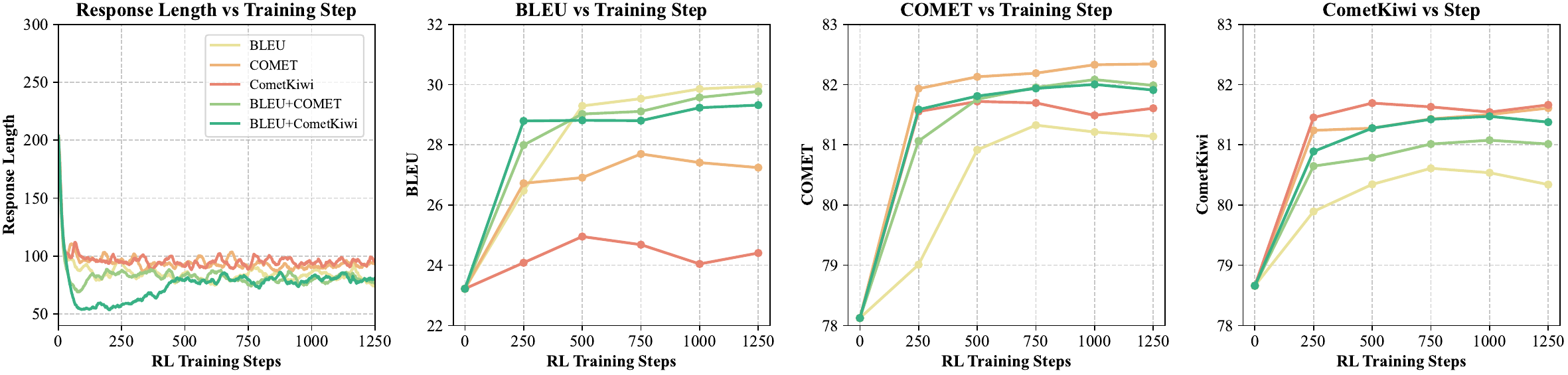}
}
\caption{Training dynamics under pure RL using different quality rewards. While translation quality improves under all settings, pure RL training fails to induce extended translation reasoning traces.}
\label{fig:zero_result}
\end{figure*}
\begin{table}[h]
    \centering
    \small
    \begin{tabularx}{\linewidth}{>{\RaggedRight}X c}
        \toprule
        \textbf{Reasoning Template} & \textbf{Rate} \\
        \midrule
        I will translate this Chinese sentence into English by identifying the key phrases and their meanings, and then constructing a coherent English sentence. & 42.66\% \\
        \midrule
        I will translate this Chinese sentence into English by identifying the key phrases and their corresponding meanings. & 24.56\% \\
        \midrule
        I will translate this Chinese sentence into English. & 5.53\% \\
        \bottomrule
    \end{tabularx}
    \caption{Top-3 reasoning template frequency.}
    \label{tab:template_analysis}
\end{table}

We first investigate the R1-Zero paradigm (Pure RL); detailed training settings are given in Appendix~\ref{appendix:r1-zero}. This setup is motivated by recent findings that RL alone can induce spontaneous reasoning capabilities in math and code tasks~\citep{deepseekai2025deepseekr1}. To test whether this emergence transfers to MDMT, we train models using GRPO with hybrid rewards. To rule out the possibility that KL regularization suppresses exploration~\citep{yeo2025demystifyinglongchainofthoughtreasoning}, we monitor token length dynamics both with and without the KL term. However, unlike the ``Aha moments'' observed in STEM tasks, our experiments reveal rapid mode collapse. 

As shown in Figures~\ref{fig:zero_w_kl} and~\ref{fig:zero_wo_kl}, reasoning traces quickly collapse into shallow patterns ($\leq 100$ tokens) regardless of the KL setting. Concretely, the model shifts toward high-frequency template recitation, suppressing diverse reasoning. As shown in Table~\ref{tab:template_analysis}, the top three templates account for about 73\% of all generated traces in the Zh$\rightarrow$En direction. This degeneration reveals a key misalignment: MDMT requires domain-aware reasoning, yet without proper initialization, the model produces reasoning that is too brief and overly templated to be elicited reliably. To address this, our cold-start phase explicitly initializes adaptive reasoning behavior before RL.

\subsection{Reasoning Challenges}
\label{sec:sft_rl_challenge}

\begin{table}[t]
\centering
\small
\setlength{\tabcolsep}{2pt}
\begin{tabular*}{\columnwidth}{@{\extracolsep{\fill}} l ccc ccc }
\toprule
\multirow{2}{*}{\textbf{Model}} & \multicolumn{3}{c}{\textbf{Easy}} & \multicolumn{3}{c}{\textbf{Hard}} \\
\cmidrule(lr){2-4} \cmidrule(lr){5-7}
 & \textbf{Quality} & \textbf{Token} & \textbf{Time} & \textbf{Quality} & \textbf{Token} & \textbf{Time} \\
\midrule
\multicolumn{7}{c}{\cellcolor{gray!15}\textbf{\textit{In-Domain}}} \\
SFT-Parallel & 64.85 & 13 & 12 & 61.72 & 40 & 15 \\
General-CoT & 61.54 & 311 & 72 & 56.53 & 551 & 80 \\
Domain-CoT & 67.36 & 558 & 136 & 62.65 & 883 & 149 \\
\midrule
\multicolumn{7}{c}{\cellcolor{gray!15}\textbf{\textit{Out-of-Domain}}} \\
SFT-Parallel & 67.52 & 13 & 7 & 59.11 & 36 & 8 \\
General-CoT & 66.59 & 354 & 96 & 60.55 & 517 & 97 \\
Domain-CoT & 69.81 & 634 & 125 & 62.52 & 840 & 138 \\
\bottomrule
\end{tabular*}
\caption{Performance by difficulty for SFT-Parallel, General-CoT, and Domain-CoT. \textit{Quality} is computed as the average of BLEU, COMET, and COMETKIWI. \textit{Time} denotes latency in milliseconds.}
\label{tab:cognitive_redundancy}
\end{table}

To evaluate the trade-off between reasoning depth and computational cost, we compare three Qwen2.5-7B-Instruct variants: SFT-Parallel (System~1), General-CoT (System~2), and Domain-CoT, which extends General-CoT with a domain-aware prompt. Details are given in Appendix~\ref{appendix:r1-cold-start}.

\paragraph{Lack of Domain Awareness.}
As shown in Table~\ref{tab:cognitive_redundancy}, General-CoT performs poorly on in-domain data because its reasoning lacks explicit domain grounding, often defaulting to generic translations rather than domain-specific terminology and fixed expressions. 
By contrast, SFT-Parallel performs well in these cases by matching the distributional patterns of its training data. 
This same contrast also explains why General-CoT can be more competitive on OOD data, particularly on harder samples: when the input does not closely match the domain patterns seen in training, intermediate reasoning helps the model better handle syntax, ambiguity, and contextual inference than standard parallel SFT. 
Importantly, adding explicit domain reasoning largely restores in-domain quality, improving over General-CoT from 61.54 to 67.36 on Easy samples and from 56.53 to 62.65 on Hard samples. This confirms that lack of domain awareness is a major source of in-domain degradation.

\paragraph{Reasoning Redundancy.}
However, domain awareness alone is not sufficient. 
General-CoT applies essentially the same reasoning strategy regardless of input difficulty, resulting in substantial redundancy.  
On Easy samples, this redundancy is clearly detrimental: despite using 23.9$\times$ more tokens in-domain (311 vs.\ 13) and 27.2$\times$ more on OOD data (354 vs.\ 13), it still underperforms SFT-Parallel by 3.31 and 0.93 quality points, respectively.
Domain-CoT does not solve this problem. Although it restores in-domain quality, it further increases token usage to 42.9$\times$ on in-domain Easy samples (558 vs.\ 13) and 48.8$\times$ on OOD Easy samples (634 vs.\ 13). 
Similar patterns also hold on Hard samples. 
These results show that prompt-level domain awareness alone cannot resolve overthinking, motivating \textbf{\texttt{TwT}}, which adaptively shifts between System~1 and System~2 to jointly address domain sensitivity and reasoning efficiency.

\section{Method}
\label{sec:method}
\begin{figure*}[th]
    \centering
    \includegraphics[width=0.9\textwidth]{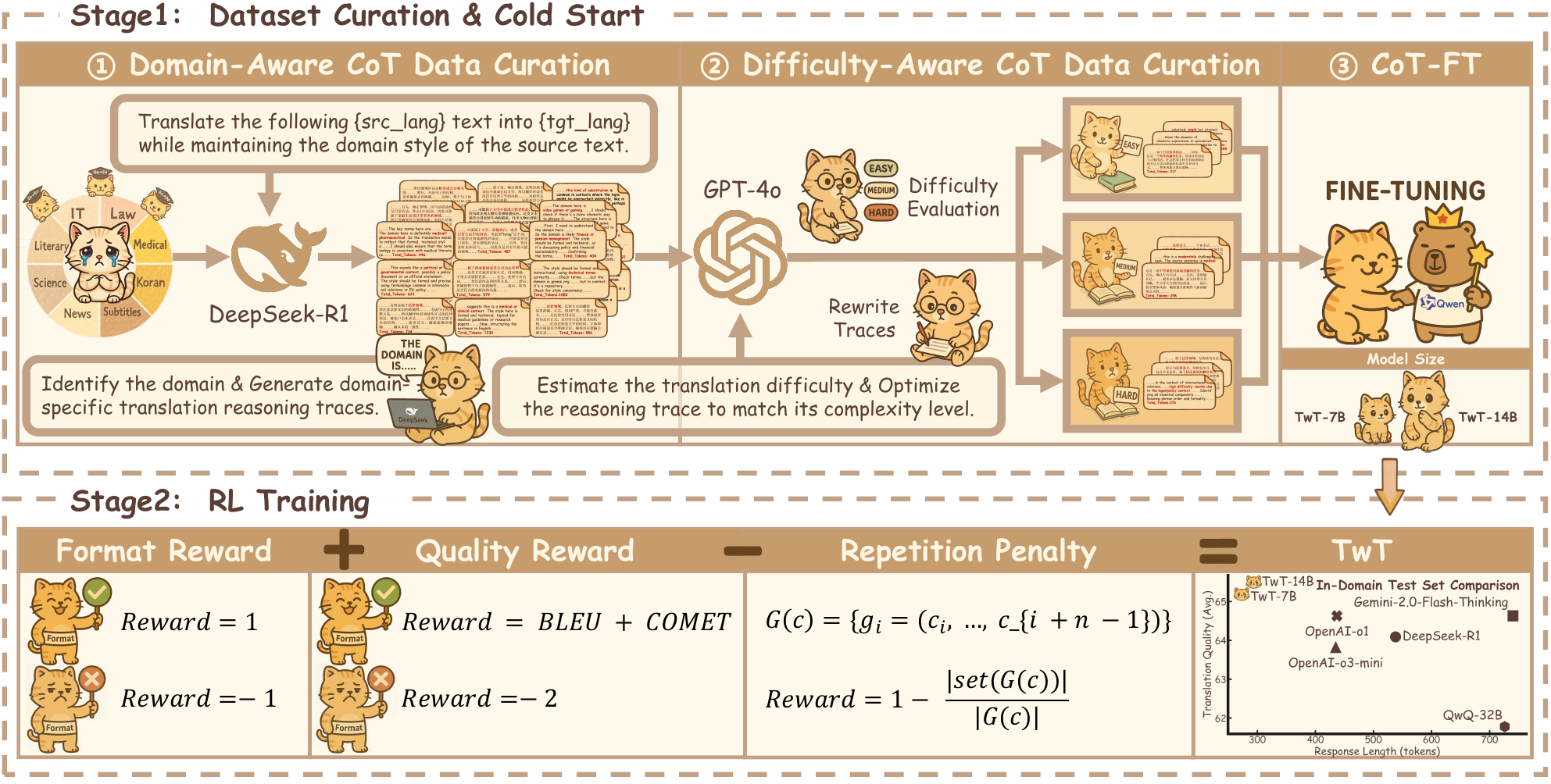}
    \caption{Overview of \textbf{\texttt{TwT}} training pipeline. \textbf{\texttt{TwT}} is first fine-tuned on difficulty-adaptive Long CoT traces distilled from DeepSeek-R1 and rewritten by GPT-4o for cognitive alignment. RL is then applied with a hybrid reward.}
    \label{fig:method}
\end{figure*}

We propose \textbf{\texttt{TwT}}, a resource-rational approach that dynamically allocates reasoning effort by input difficulty, mimicking human translation process. As shown in Figure~\ref{fig:method}, training proceeds in two stages.

\subsection{Cold Start}
\label{sec:cold_start}

To align the model's reasoning behavior with the human translation process, we construct a Difficulty-Adaptive CoT Dataset that equips the backbone LLM with adaptive reasoning capabilities. The dataset is curated through a multi-agent distillation-adaptation pipeline. We first prompt DeepSeek-R1 with domain-aware instructions to generate high-quality CoT traces tailored to different domains. These serve as the initial reasoning demonstrations. We then employ GPT-4o, which is verified to align best with human judgment (Appendix~\ref{apd:difficulty_reliability}), to assess input difficulty based on the theory of Rich Points. Specifically, difficulty is defined along four linguistic dimensions: sentence complexity, vocabulary rarity, grammatical divergence, and contextual nuance.
The corresponding evaluation prompt is shown in Figures~\ref{fig:difficulty_polish_prompt} and~\ref{fig:difficulty_eval}.

Conditioned on this assessment, GPT-4o rewrites raw traces into adaptive formats: Easy inputs are reformulated into concise System~1 checks to minimize token usage, whereas Hard inputs retain comprehensive System~2 deliberations for structural and terminological verification. We then perform SFT on the backbone LLM using this compact dataset ($\sim$7k examples). We define reasoning depth as the length of the generated reasoning trace (i.e., number of tokens). This process establishes an initial policy distribution over difficulty-aware reasoning strategies, enabling the model to autonomously modulate its reasoning depth. Representative examples are shown in Figures~\ref{fig:example_easy_mid} and~\ref{fig:example_hard}. The resulting data efficiency makes our approach particularly suitable for low-resource settings, demonstrating that robust adaptive reasoning can be achieved with a modest CoT-SFT seed dataset.

\subsection{RL Training}
\label{sec:rl}
To scale this adaptive reasoning behavior, we adopt the GRPO algorithm with hybrid quality rewards, which serve as outcome-driven constraints that encourage the model to identify Rich Points and allocate deep reasoning selectively—only where it leads to measurable quality improvements. By mitigating indiscriminate overthinking (Section~\ref{sec:sft_rl_challenge}), we improve reasoning efficiency and align the model with the resource-rational principle introduced in Section~\ref{sec:intro}.
The final reward $r$ is crafted from three components to ensure alignment:
$$r = r_{f} + r_{q} - \lambda \cdot r_{\text{rep}}$$

\paragraph{Format Reward ($r_f$).}
We employ a binary format reward ($r_f \in \{1, -1\}$) to strictly enforce the reasoning-translation structure defined in Figure~\ref{fig:prompt_zero}.

\paragraph{Hybrid Quality Reward ($r_q$).}
To prevent the model from producing plausible but functionally ineffective reasoning, we introduce a verifiable hybrid feedback signal.
Building upon our preliminary analysis in Section~\ref{sec:pure_rl_challenge}, we identified a critical reward hacking phenomenon. As illustrated by the red curve in Figure~\ref{fig:zero_w_kl}, optimizing solely for a semantic metric (CometKiwi/COMET) leads to a significant metric divergence: despite high semantic reward scores, the lexical accuracy (BLEU) degrades rapidly during training.
This confirms that the model hacks the reward by generating vague paraphrases or copying source tokens to maximize semantic similarity, effectively abandoning translation fidelity.
To remedy this and strictly enforce alignment, we adopt a hybrid quality reward that combines BLEU (\text{B}) with COMET (\text{C}):
$$r_{q}= \begin{cases}\text{B}(\mathbf{\hat{y}}, \mathbf{y}) + \text{C}(\mathbf{x}, \mathbf{\hat{y}}, \mathbf{y}) & \text {if \ } r_f=1 \\ -2 & \text {if \ }  r_f = -1 \end{cases}$$

This design effectively stabilizes the optimization process. As evidenced by the training dynamics of our \textbf{\texttt{TwT}} models (Figure~\ref{fig:traing_dynamics}), both the 7B and 14B models (Figures~\ref{fig:curves_7B} and~\ref{fig:curves_14B}) exhibit synchronous improvements in lexical and semantic metrics without divergence, validating that the hybrid signal successfully grounds the reasoning process in accurate translation outcomes.

\paragraph{N-gram Repetition Penalty ($r_{\text{rep}}$).}
We penalize redundant reasoning to discourage degenerate loops and promote efficient token usage. For a given reasoning trace $\mathbf{c}$, let $\mathbf{G}(\mathbf{c})$ denote the list of contiguous $n$-grams. In our experiments, we set $n=20$. We compute the ratio of repeated $n$-grams to discourage repetitive, loop-like patterns:
$$r_{\text{rep}} =1 - \frac{\lvert \text{set}\left(\mathbf{G}(\mathbf{c})\right) \rvert}{\lvert \mathbf{G}(\mathbf{c}) \rvert} \in [0,1]$$

\section{Experiments}

\begin{table*}[!ht]
\centering
\resizebox{\textwidth}{!}{%
\begin{tabular}{lccccccccccccccccccccc}
\toprule
\multirow{2}{*}{\textbf{Method}} & \multicolumn{2}{c}{\textbf{Laws}} & \multicolumn{2}{c}{\textbf{News}} & \multicolumn{2}{c}{\textbf{Science}} & \multicolumn{2}{c}{\textbf{Subtitles}} & \multicolumn{2}{c}{\textbf{Literary}} & \multicolumn{2}{c}{\textbf{IT}} & \multicolumn{2}{c}{\textbf{Koran}} & \multicolumn{2}{c}{\textbf{Medical}} & \multicolumn{2}{c}{\textbf{Average}} \\

\cmidrule(lr){2-3} \cmidrule(lr){4-5} \cmidrule(lr){6-7} \cmidrule(lr){8-9} \cmidrule(lr){10-11} \cmidrule(lr){12-13} \cmidrule(lr){14-15} \cmidrule(lr){16-17} \cmidrule(lr){18-19}

 & \textbf{Quality} & \textbf{Tokens} & \textbf{Quality} & \textbf{Tokens} & \textbf{Quality} & \textbf{Tokens} & \textbf{Quality} & \textbf{Tokens} & \textbf{Quality} & \textbf{Tokens} & \textbf{Quality} & \textbf{Tokens} & \textbf{Quality} & \textbf{Tokens} & \textbf{Quality} & \textbf{Tokens} & \textbf{Quality} & \textbf{Tokens} \\
\midrule

\rowcolor{gray!15}
\multicolumn{19}{c}{\textbf{\textit{Large Language Models}}} \\
DeepSeek-V3 & \textbf{77.72} & - & \textbf{69.41} & - & \textbf{69.09} & - & 62.93 & - & 56.83 & - & 66.97 & - & 57.73 & - & 69.16 & - & 66.23 & - \\
Gemini-2.0-Flash & 76.57 & - & {\ul 69.30} & - & {\ul 68.80} & - & 62.96 & - & 57.41 & - & 66.54 & - & 58.13 & - & 70.21 & - & 66.24 & - \\
GPT-4o & 73.77 & - & 68.61 & - & 68.09 & - & 62.92 & - & 57.47 & - & 66.37 & - & 57.78 & - & 69.34 & - & 65.54 & - \\

\rowcolor{gray!15}
\multicolumn{19}{c}{\textbf{\textit{Large Reasoning Models}}} \\
DeepSeek-R1 & \textbf{77.72} & 577 & 68.47 & 498 & 68.46 & 478 & 61.81 & 514 & 53.95 & 574 & 66.28 & 593 & 57.48 & 790 & 69.01 & 667 & 65.40 & 586 \\
Gemini-2.0-Flash-Thinking & 76.21 & 702 & 68.24 & 1149 & 68.17 & 1092 & 61.98 & 708 & 57.23 & 781 & 66.20 & 345 & 58.13 & 677 & 69.73 & 415 & 65.74 & 734 \\
OpenAI-o3-mini & 71.61 & 428 & 68.18 & 443 & 68.01 & 385 & 62.42 & 355 & 56.97 & 546 & 65.99 & 343 & 56.42 & 511 & 68.40 & 346 & 64.75 & 420 \\
OpenAI-o1 & 73.85 & 478 & 68.67 & 408 & 68.60 & 367 & 62.62 & 340 & 57.28 & 521 & 66.45 & 403 & 57.94 & 506 & 69.20 & 441 & 65.58 & 433 \\
GPT-5 & 76.28 & 784 & 69.07 & 740 & 68.58 & 606 & 62.74 & 519 & 56.27 & 859 & 66.57 & 492 & 58.71 & 751 & 70.14 & 531 & 66.05 & 660 \\
QwQ-32B & 71.84 & 667 & 67.88 & 584 & 67.86 & 563 & 61.80 & 584 & 55.09 & 863 & 61.44 & 583 & 55.81 & 963 & 66.68 & 735 & 63.55 & 693 \\

\rowcolor{gray!15}
\multicolumn{19}{c}{\textbf{\textit{MT-Specialized Models}}} \\
SFT-Parallel-7B & 76.58 & - & 66.08 & - & 66.38 & - & 62.87 & - & 55.96 & - & 68.02 & - & 58.05 & - & 70.21 & - & 65.52 & - \\
ALMA-7B-R & 67.88 & - & 63.37 & - & 62.77 & - & 59.38 & - & 54.52 & - & 64.20 & - & 55.04 & - & 67.46 & - & 61.83 & - \\
ALMA-13B-R & 70.11 & - & 64.65 & - & 64.23 & - & 60.22 & - & 55.30 & - & 64.54 & - & 55.71 & - & 68.40 & - & 62.89 & - \\
TowerInstruct-7B-v0.2 & 73.91 & - & 65.93 & - & 65.45 & - & 61.07 & - & 55.32 & - & 66.78 & - & 50.05 & - & {\ul 70.73} & - & 63.66 & - \\
TowerInstruct-13B-v0.1 & 74.65 & - & 66.90 & - & 66.20 & - & 61.89 & - & 56.09 & - & 67.18 & - & 49.93 & - & \textbf{71.49} & - & 64.29 & - \\
CoT-FT-7B & {\ul 76.72} & 51 & 66.24 & 42 & 66.07 & 39 & 62.76 & 29 & 55.57 & 52 & 67.82 & 35 & 57.57 & 45 & 70.38 & 46 & 65.39 & 42 \\
MT-R1-Zero-7B & 68.93 & 72 & 67.41 & 64 & 66.98 & 61 & 61.97 & 55 & 55.80 & 69 & 65.60 & 56 & 55.52 & 71 & 64.43 & 71 & 63.33 & 65 \\
SSR-X-Zero-7B & 69.59 & 56 & 66.00 & 52 & 66.77 & 49 & 61.89 & 39 & 55.54 & 54 & 61.22 & 36 & 55.53 & 46 & 63.99 & 50 & 62.56 & 48 \\
mExTrans-7B & 70.11 & 597 & 65.49 & 553 & 66.13 & 546 & 59.52 & 476 & 54.40 & 610 & 60.50 & 452 & 55.41 & 604 & 63.00 & 565 & 61.82 & 551 \\

\rowcolor{gray!15}
\multicolumn{19}{c}{\textbf{\textit{Our Models}}} \\
\textbf{\texttt{TwT-Qwen2.5-7B-Instruct}} & 75.35 & 310 & 68.42 & 311 & 68.24 & 294 & \textbf{63.03} & 247 & {\ul 57.86} & 281 & {\ul 68.07} & 222 & {\ul 58.92} & 269 & 70.29 & 262 & {\ul 66.27} & 274 \\
\textbf{\texttt{TwT-Qwen2.5-14B-Instruct}} & 76.58 & 320 & 68.62 & 285 & 68.32 & 272 & {\ul 62.97} & 241 & \textbf{58.14} & 354 & \textbf{68.37} & 234 & \textbf{59.35} & 336 & 70.59 & 287 & \textbf{66.62} & 291 \\
\bottomrule
\end{tabular}%
}
\caption{In-domain translation performance across eight domains, averaged over En$\rightarrow$Zh, Zh$\rightarrow$En, and De$\rightarrow$En. The \textbf{Bold} and \underline{underlined} values denote the highest and second highest scores, respectively.}
\label{tab:result_zh_en}
\end{table*}

\begin{table*}[!ht]
\centering
\setlength{\tabcolsep}{5mm}
\resizebox{\textwidth}{!}{%
\begin{tabular}{lcccccccccccccccccc}
\toprule
\multirow{2}{*}{\textbf{Method}} & \multicolumn{2}{c}{\textbf{Conversation}} & \multicolumn{2}{c}{\textbf{Ecommerce}} & \multicolumn{2}{c}{\textbf{Social}} & \multicolumn{2}{c}{\textbf{Culture}} & \multicolumn{2}{c}{\textbf{CommonSense}} & \multicolumn{2}{c}{\textbf{Average}} \\

\cmidrule(lr){2-3} \cmidrule(lr){4-5} \cmidrule(lr){6-7}  \cmidrule(lr){8-9} \cmidrule(lr){10-11}  \cmidrule(lr){12-13} 

 & \textbf{Quality} & \textbf{Tokens} & \textbf{Quality} & \textbf{Tokens} & \textbf{Quality} & \textbf{Tokens} & \textbf{Quality} & \textbf{Tokens} & \textbf{Quality} & \textbf{Tokens} & \textbf{Quality} & \textbf{Tokens} \\
\midrule

\rowcolor{gray!15}
\multicolumn{13}{c}{\textbf{\textit{Large Language Models}}} \\
DeepSeek-V3 & 68.42 & - & {\ul 66.34} & - & \textbf{66.10} & - & \textbf{69.65} & - & \textbf{65.96} & - & \textbf{67.29} & - \\
Gemini-2.0-Flash & \textbf{68.77} & - & 66.26 & - & \textbf{66.10} & - & {\ul 69.02} & - & 65.23 & - & 67.08 & - \\
GPT-4o & {\ul 68.75} & - & \textbf{66.50} & - & \textbf{66.10} & - & 69.01 & - & {\ul 65.89} & - & {\ul 67.25} & - \\

\rowcolor{gray!15}
\multicolumn{13}{c}{\textbf{\textit{Large Reasoning Models}}} \\
DeepSeek-R1 & 67.06 & 534 & 64.45 & 552 & 64.11 & 554 & 68.16 & 560 & 64.34 & 602 & 65.62 & 561 \\
Gemini-2.0-Flash-Thinking & 68.36 & 1204 & 65.83 & 822 & 65.50 & 1081 & 68.42 & 1220 & 65.44 & 2335 & 66.71 & 1332 \\
OpenAI-o3-mini & 68.04 & 290 & 65.87 & 363 & {\ul 65.58} & 372 & 67.55 & 596 & 64.45 & 436 & 66.30 & 411 \\
OpenAI-o1 & 68.47 & 327 & 65.80 & 399 & 65.38 & 405 & 67.96 & 542 & 64.78 & 392 & 66.48 & 413 \\
GPT-5 & 68.40 & 448 & 65.49 & 609 & 65.14 & 652 & 68.33 & 984 & 64.09 & 530 & 66.29 & 645 \\

\rowcolor{gray!15}
\multicolumn{13}{c}{\textbf{\textit{MT-Specialized Models}}} \\
SFT-Parallel-7B & 65.64 & - & 63.33 & - & 62.54 & - & 65.34 & - & 61.03 & - & 63.58 & - \\
ALMA-7B-R & 64.71 & - & 62.54 & - & 62.68 & - & 66.63 & - & 62.02 & - & 63.71 & - \\
ALMA-13B-R & 66.03 & - & 63.37 & - & 63.51 & - & 60.94 & - & 62.91 & - & 63.35 & - \\
CoT-FT-7B & 65.43 & 31 & 63.26 & 45 & 62.06 & 42 & 64.66 & 54 & 61.08 & 33 & 63.30 & 41 \\
MT-R1-Zero-7B & 66.59 & 53 & 64.16 & 66 & 63.69 & 65 & 66.23 & 79 & 62.32 & 51 & 64.60 & 63 \\
SSR-X-Zero-7B & 65.70 & 37 & 63.46 & 50 & 63.31 & 49 & 64.35 & 66 & 62.18 & 34 & 63.80 & 47 \\
mExTrans-7B & 63.47 & 464 & 61.74 & 566 & 61.20 & 555 & 65.11 & 631 & 59.92 & 470 & 62.29 & 537 \\

\rowcolor{gray!15}
\multicolumn{13}{c}{\textbf{\textit{Our Models}}} \\
\textbf{\texttt{TwT-Qwen2.5-7B-Instruct}} & 67.72 & 231 & 65.71 & 273 & 65.48 & 269 & 67.82 & 352 & 64.48 & 219 & 66.25 & 269 \\
\textbf{\texttt{TwT-Qwen2.5-14B-Instruct}} & 67.77 & 240 & 65.79 & 309 & 65.47 & 298 & 68.55 & 333 & 64.72 & 259 & 66.46 & 288 \\
\bottomrule
\end{tabular}%
}
\caption{OOD translation performance across five domains, averaged over En$\rightarrow$Zh, Zh$\rightarrow$En, and De$\rightarrow$En.}
\label{tab:result_ood}
\end{table*}

\begin{table*}[t]
\centering
\resizebox{\textwidth}{!}{%
\begin{tabular}{lcccccccccccc}
\toprule

\multirow{2}{*}{\textbf{Method}} & \multicolumn{2}{c}{\textbf{En$\rightarrow$Zh}} & \multicolumn{2}{c}{\textbf{Zh$\rightarrow$En}} & \multicolumn{2}{c}{\textbf{De$\rightarrow$En}} & \multicolumn{2}{c}{\textbf{En$\rightarrow$X}} & \multicolumn{2}{c}{\textbf{X$\rightarrow$En}} & \multicolumn{2}{c}{\textbf{Average}} \\

\cmidrule(lr){2-3} \cmidrule(lr){4-5} \cmidrule(lr){6-7}  \cmidrule(lr){8-9} \cmidrule(lr){10-11}  \cmidrule(lr){12-13} 

 & \textbf{Quality} & \textbf{Tokens} & \textbf{Quality} & \textbf{Tokens} & \textbf{Quality} & \textbf{Tokens} & \textbf{Quality} & \textbf{Tokens} & \textbf{Quality} & \textbf{Tokens} & \textbf{Quality} & \textbf{Tokens} \\
\midrule
\rowcolor{gray!15}
\multicolumn{13}{c}{\textbf{\textit{Large Language Models}}} \\
 
Qwen2.5-7B-Instruct & 67.87 & - & 50.91 & - & 62.90 & - & 37.61 & - & 56.99 & - & 55.25 & - \\
Gemma-2-9B-IT & 66.50 & - & 52.64 & - & 60.64 & - & {\ul 54.28} & - & {\ul 66.38} & - & 60.09 & - \\

\rowcolor{gray!15}
\multicolumn{13}{c}{\textbf{\textit{MT-Specialized Models}}} \\

ALMA-7B-R & 64.08 & - & 54.52 & - & 62.89 & - & 44.68 & - & 44.71 & - & 54.18 & - \\
Tower-Plus-9B & {\ul 69.85} & - & 57.41 & - & \textbf{67.04} & - & 46.46 & - & 63.02 & - & {\ul 60.76} & - \\
SFT-Parallel-7B & 69.44 & - & 55.96 & - & 66.15 & - & 32.44 & - & 55.70 & - & 55.94 & - \\
MT-R1-Zero-7B & 67.52 & 62 & 55.80 & 69 & 62.56 & 66 & 40.98 & 358 & 58.33 & 74 & 57.04 & 126 \\
SSR-X-Zero-7B & 68.55 & 50 & 56.52 & 54 & 62.86 & 45 & 40.43 & 306 & 58.19 & 48 & 57.31 & 101 \\
mExTrans-7B & 66.70 & 537 & 54.40 & 610 & 60.74 & 544 & 42.72 & 1047 & 55.99 & 731 & 56.11 & 694 \\

\rowcolor{gray!15}
\multicolumn{13}{c}{\textbf{\textit{Our Models}}} \\

\textbf{\texttt{TwT-Qwen2.5-7B-Instruct}} & \textbf{69.99} & 298 & {\ul 57.86} & 281 & 66.45 & 256 & 41.13 & 483 & 58.84 & 328 & 58.85 & 329 \\
\textbf{\texttt{TwT-Gemma-2-9B-IT}} & 69.07 & 227 & \textbf{58.12} & 249 & {\ul 66.95} & 218 & \textbf{54.65} & 280 & \textbf{67.04} & 257 & \textbf{63.17} & 246 \\
\bottomrule
\end{tabular}%
}
\caption{Results on seen and unseen language directions. En, Zh, and De are \emph{seen} languages, while X denotes \emph{unseen} languages; En$\rightarrow$X and X$\rightarrow$En report averages over English$\leftrightarrow$unseen-language directions.}

\label{tab:result_lang}
\end{table*}

\begin{table}[t]
\centering
\resizebox{\linewidth}{!}{%
\begin{tabular}{lcccc}
\toprule
\multirow{2}{*}{Method} & \multicolumn{2}{c}{In-Domain} & \multicolumn{2}{c}{Out-of-Domain} \\
\cmidrule(lr){2-3} \cmidrule(lr){4-5}
 & Quality & Tokens & Quality & Tokens \\
 \cmidrule(lr){1-5}
\textbf{\texttt{TwT-Qwen2.5-7B-Instruct}} & 64.77 & 278 & 66.28 & 263 \\
w/o RP & 64.52 & 301 & 66.12 & 271 \\
w/o RP + w/o Adaptive CoT & 64.68 & 748 & 66.17 & 720 \\
w/o RP + w/o Cold Start & 64.18 & 62 & 65.29 & 59 \\
SFT only w/ Adaptive CoT & 62.47 & 253 & 64.09 & 219 \\
SFT only w/ domain-aware CoT & 61.65 & 479 & 63.47 & 452 \\
SFT only w/ general CoT & 61.39 & 531 & 63.11 & 519 \\
\bottomrule
\toprule
\rowcolor{mygray}
\multicolumn{5}{c}{\textbf{\textit{Backbone Models}}} \\
Qwen2.5-7B-Instruct & 61.99 & - & 64.71 & - \\
Llama-3.1-8B-Instruct & 59.84 & - & 63.33 & - \\
Gemma-2-9B-IT & 59.93 & - & 63.68 & - \\

\rowcolor{mygray}
\multicolumn{5}{c}{\textbf{\textit{Our Models}}} \\
\textbf{\texttt{TwT-Qwen2.5-7B-Instruct}} & 64.77 & 278 & 66.28 & 263 \\
\textbf{\texttt{TwT-Llama-3.1-8B-Instruct}} & 63.20 & 305 & 64.44 & 298 \\
\textbf{\texttt{TwT-Gemma-2-9B-IT}} & 64.71 & 231 & 66.05 & 220 \\
\bottomrule

\end{tabular}%
}
\caption{Ablation study on in-domain and OOD translation test sets. Results are averaged at the dataset level for each setting. RP = repetition penalty.}

\label{tab:ablation}
\end{table}

\subsection{Experimental Settings}

\paragraph{Dataset.} 
We use two datasets for training: (1) a curated set of 7K difficulty-adaptive Long CoT examples spanning 10 domains and three translation directions (De$\rightarrow$En, En$\rightarrow$Zh, Zh$\rightarrow$En) for cold-start SFT, and (2) a separate 20K-sample dataset for RL, constructed from multi-domain parallel corpora.
For evaluation, we adopt both in-domain test sets and diverse OOD benchmarks, and additionally include multilingual test sets covering both seen and unseen language pairs.
Full dataset details are provided in Appendix~\ref{apd:dataset}.

\paragraph{Implementation Details.}
For cold start, we use LLaMA-Factory\footnote{\url{https://github.com/hiyouga/LLaMA-Factory}}~\cite{zheng-etal-2024-llamafactory} with Qwen2.5-7B-Instruct, Qwen2.5-14B-Instruct, and Gemma-2-9B-IT as backbones. We train them on the 7K difficulty-adaptive Long CoT examples for 1 epoch with full-parameter optimization on 8 NVIDIA A100 80GB GPUs, using AdamW with a learning rate of $1\mathrm{e}{-5}$, a total batch size of 32, a cosine learning rate scheduler, a warm-up ratio of 0.1, a maximum input sequence length of 4096, and DeepSpeed ZeRO Stage 3. The cold-start stage completes within 10 minutes.
For RL, we use \texttt{verl}\footnote{\url{https://github.com/volcengine/verl}}~\cite{sheng2025hybridflow} and train for 1 epoch on 8 NVIDIA A100 80GB GPUs with a total batch size of 16, rollout number 16, rollout temperature 1.0, learning rate $1\mathrm{e}{-6}$, KL loss coefficient $\beta = 1\mathrm{e}{-3}$, maximum response length 2048, and repetition penalty with $n=20$. RL training takes about 10 hours. During inference, we use \texttt{vLLM}\footnote{\url{https://github.com/vllm-project/vllm}}~\citep{woosuk2025vllm} for efficient decoding with temperature 0.0 and repetition penalty 1.05.

\paragraph{Metrics.} 
We report \textbf{\textit{Quality}}, defined as the average of BLEU, COMET~\citep{rei-etal-2020-comet}, and CometKiwi~\citep{rei-etal-2022-cometkiwi}, and \textbf{\textit{Tokens}}, the average length of the generated CoT.
The full metric breakdowns are provided in Appendix~\ref{appendix:metric_breakdown}.

\paragraph{Baselines.}
We compare our method against three categories of models:  
(1) \textbf{\textit{general-purpose LLMs}} such as DeepSeek-V3~\cite{deepseekai2025deepseekv3technicalreport}, Gemini-2.0-Flash~\cite{googledeepmind2024gemini20}, GPT-4o~\cite{openai2024gpt4ocard}, and open-source models like LLaMA3.1-8B-Instruct~\cite{grattafiori2024llama3herdmodels}, Gemma-2-9B-IT~\cite{gemmateam2024gemma2improvingopen}, and Qwen2.5 series (7B, 14B, 32B)~\cite{qwen2025qwen25}.  
(2) \textbf{\textit{reasoning-oriented LRMs}}, such as DeepSeek-R1~\cite{deepseekai2025deepseekr1}, Gemini-2.0-Flash-Thinking~\cite{googledeepmind2024gemini20flashthinking}, OpenAI o1~\cite{jaech2024openai}, o3-mini~\cite{openai2025o3mini}, GPT-5~\cite{openai_gpt5}, and QwQ-32B~\cite{qwq32b};  
(3) \textbf{\textit{MT-specialized models}}, including non-reasoning LLMs such as TowerInstruct~\cite{tower_llm_2024}, ALMA-R~\cite{xu2024a, xu2024contrastive}, SFT-Parallel (Qwen2.5-7B-Instruct fine-tuned on 27K parallel pairs), CoT-FT~\cite{hu-etal-2024-large-language}, and Tower-Plus-9B~\cite{rei2025towerbridginggeneralitytranslation}; with reasoning-oriented models including MT-R1-Zero~\cite{feng2025mtr1zeroadvancingllmbasedmachine}, mExTrans~\cite{wang2025extransmultilingualdeepreasoning}, and SSR-X-Zero~\cite{yang2025ssrzerosimpleselfrewardingreinforcement}. For fairer comparison, Appendix~\ref{apd:mt_data_scale} details the training data size of MT-specialized baselines.

\subsection{Cross-Domain Generalization}
\label{sec:domain_results}

\paragraph{In-Domain.} 
As shown in Table~\ref{tab:result_zh_en}, \textbf{\texttt{TwT-14B}} achieves a SOTA average quality score of 66.62, outperforming both strong open-source LRMs such as DeepSeek-R1 (65.40) and closed-source models including GPT-5 (66.05). Notably, our smaller variant, \textbf{\texttt{TwT-7B}}, also surpasses dedicated MT systems, demonstrating that our method scales effectively with model size.
Compared to mExTrans-7B, which is also trained under the R1 paradigm, \textbf{\texttt{TwT-7B}} yields a substantial improvement of $+$4.45 points while reducing reasoning overhead by 50.27\%, highlighting the superior efficiency of our difficulty-adaptive mechanism.
The benefits of adaptive reasoning are especially evident in structurally complex and culturally nuanced domains. In the Literary domain, \textbf{\texttt{TwT-14B}} consistently outperforms three representative paradigms: it exceeds the System~1 baseline SFT-Parallel by $+$2.18 points, the heavy-reasoning System~2 model DeepSeek-R1 by $+$4.19 points, and the pure RL-based MT-R1-Zero-7B by $+$2.34 points. These results suggest that, for multi-domain translation, neither shallow System~1 execution, indiscriminate System~2 overthinking, nor unguided pure RL exploration alone yields optimal performance. In contrast, our model autonomously modulates reasoning depth to strike a more effective balance between literal accuracy and stylistic adequacy.

\paragraph{OOD.}
\textbf{\texttt{TwT-7B}} achieves a strong average score of 66.25 on five OOD test sets, surpassing MT-specialized baselines such as SFT-Parallel-7B (63.58), which exhibit poor generalization under domain shift.
Moreover, \textbf{\texttt{TwT-7B}} exhibits a favorable quality-efficiency trade-off, surpassing DeepSeek-R1 while reducing reasoning overhead by 292 tokens. Even in unfamiliar domains, it avoids excessive deliberation by leveraging compact, internalized translation procedures. These results indicate that \textbf{\texttt{TwT}} captures domain-agnostic translation logic rather than relying on domain-specific memorization, approaching the SOTA performance of GPT-4o at a fraction of the computational cost and model size.

\subsection{Multilingual Generalization}
\label{sec:lang_results}

On seen directions, \textbf{\texttt{TwT-7B}} improves Zh$\rightarrow$En performance by $+$6.95 over its base model and $+$1.90 over SFT-Parallel-7B, under the same training data. On unseen directions (En$\rightarrow$X)\footnote{En$\rightarrow$X and X$\rightarrow$En are averaged over 59 unseen languages from FLORES+; see Appendix~\ref{apd:multilingual} for the full list.}, \textbf{\texttt{TwT-Gemma-2-9B-IT}}, trained on only 27K examples, outperforms the multilingual Tower-Plus-9B, built on the same backbone but trained on 286K examples, by a substantial margin of $+$8.19. Overall, \textbf{\texttt{TwT}} achieves the highest average score across all directions, indicating that its reasoning mechanism generalizes beyond language boundaries and captures transferable alignment strategies.

\subsection{Ablation Study}
\label{sec:ablation}
We conduct a comprehensive ablation study to assess the contribution of each component in the \textbf{\texttt{TwT}} framework and validate its generalizability across backbone architectures (Table~\ref{tab:ablation}).
Removing the repetition penalty (w/o RP) leads to a slight quality drop and longer outputs, indicating its role as a regularizer rather than a performance driver. In contrast, removing both the repetition penalty and the difficulty-adaptive rewriting (w/o RP + w/o Adaptive CoT) results in comparable quality but significantly increases reasoning length (from 278 to 748 tokens), highlighting the critical role of adaptive rewriting in controlling verbosity and ensuring inference efficiency.
We further examine the necessity of the two-stage training pipeline. Eliminating the cold-start SFT phase (w/o RP + w/o Cold Start) causes performance degradation and length collapse (to 62 tokens), suggesting that RL alone fails to induce structured reasoning behavior. To isolate the impact of SFT data quality, we compare three variants: difficulty-adaptive CoT yields the best result (62.47), followed by domain-aware (61.65) and general CoT (61.39), showing a clear performance hierarchy. Still, only the full \textbf{\texttt{TwT}} pipeline achieves the highest score (64.77), confirming that RL is indispensable for turning the adaptive reasoning patterns from mere imitation into an internalized and optimized translation strategy.
Lastly, we apply \textbf{\texttt{TwT}} to three backbone models: Qwen2.5-7B, Llama-3.1-8B, and Gemma-2-9B, and observe consistent improvements. For instance, \textbf{\texttt{TwT}} improves Gemma-2-9B-IT from 59.93 to 64.71 in-domain and from 63.68 to 66.05 OOD, demonstrating that \textbf{\texttt{TwT}} is a model-agnostic framework that robustly enhances translation reasoning regardless of the underlying architecture.

\section{Empirical Analysis}

\subsection{Human Reasoning Alignment}
\label{sec:human_comparison}
To benchmark \textbf{\texttt{TwT}}'s reasoning against human cognition, we conduct a qualitative analysis on 10 Zh$\rightarrow$En examples, with expert commentary from a translation studies faculty member. A representative case is shown in Appendix~\ref{apd:human_eval_case}.

\paragraph{Cognitive Convergence.}
The analysis revealed that \textbf{\texttt{TwT}}'s reasoning exhibits strong parallels with human translators in early-stage decision-making. Specifically, \textbf{\texttt{TwT}} effectively (1) identifies translation domain and stylistic register, (2) handles complex sentence structures with appropriate syntactic parsing, and (3) demonstrates context-aware terminology adaptation. For example, it consistently distinguishes between literary and technical expressions and adjusts lexical choices accordingly. Its structured CoT mirrors key aspects of professional reasoning—such as coherence maintenance, discourse flow control, and sensitivity to stylistic norms—indicating that \textbf{\texttt{TwT}} has internalized domain-aware reasoning behavior resembling human translation logic.

\paragraph{Pragmatic Divergence.}
Despite these strengths, \textbf{\texttt{TwT}} still shows gaps compared with expert translators. It occasionally struggles with cross-sentence consistency in terminology, especially when handling long-form repetitions or abbreviated references. Moreover, its output lacks fine-grained control over tone, idiomaticity, and cultural adaptation, which human translators adjust based on pragmatic context and target audience. These issues suggest that \textbf{\texttt{TwT}}'s reasoning remains less flexible in discourse-level adaptation, reflecting the absence of high-level pragmatic awareness. Future work seeks to bridge these gaps by integrating process-oriented feedback, thereby fostering deeper pragmatic alignment with human cognitive processes.

\subsection{Reasoning Redundancy Reduction}
\label{sec:efficiency_eval}
\begin{table}[t]
    \setlength{\tabcolsep}{15pt}
    \centering
    \resizebox{\linewidth}{!}{%
        \begin{tabular}{lcc}
            \toprule
            \textbf{Redundancy Type} & \textbf{Resolved (\%)} & \textbf{Not Resolved (\%)} \\
            \midrule
            Over-segmentation & 87.3 & 12.7 \\
            Unnecessary linguistic explanation & 93.2 & 6.8 \\
            Semantic repetition & 96.9 & 3.1 \\
            Irrelevant information & 93.7 & 6.3 \\
            Redundant alternative translations & 96.5 & 3.5 \\
            Low-density long descriptions & 97.2 & 2.8 \\
            \bottomrule
        \end{tabular}%
    }
    \caption{Resolution rates of six redundancies by \textbf{\texttt{TwT}}.}
    \label{tab:redundancy_eval}
\end{table}

To assess whether \textbf{\texttt{TwT}} eliminates unnecessary computation, we employ DeepSeek-V3.2 to detect six distinct forms of reasoning redundancy across 15 domains, utilizing the prompt provided in Figure~\ref{fig:redundancy_prompt}.
As shown in Table~\ref{tab:redundancy_eval}, \textbf{\texttt{TwT}} demonstrates exceptional efficiency, successfully resolving over 94\% of the redundant steps observed in the SFT-with-RL baseline described in Section~\ref{sec:sft_rl_challenge}. 
Notably, it achieves a 97.2\% resolution rate for Low-Density Long Descriptions, verifying its ability to compress verbose reasoning into high-density insights, while maintaining 87.3\% resolution for structural issues like Over-Segmentation. 
These results confirm that the model has internalized resource-rational adaptive reasoning behavior, effectively activating System~2 reasoning for Rich Points while avoiding unnecessary elaboration on straightforward segments.

\subsection{Translation Difficulty Adaptation}

\begin{table}[t]
\centering
\resizebox{\linewidth}{!}{%
\begin{tabular}{lcccccccc}
\toprule
\multirow{2}{*}{Method} & \multicolumn{2}{c}{Easy} & \multicolumn{2}{c}{Medium} & \multicolumn{2}{c}{Hard} & \multicolumn{2}{c}{All} \\
\cmidrule(lr){2-3} \cmidrule(lr){4-5} \cmidrule(lr){6-7} \cmidrule(lr){8-9}
 & Quality & Tokens & Quality & Tokens & Quality & Tokens & Quality & Tokens \\
\cmidrule(lr){1-9}
DeepSeek-R1 & 66.30 & 480 & 63.82 & 556 & 62.15 & 579 & 64.09 & 538 \\
Gemini-2.0-Flash-Thinking & 66.62 & 432 & 64.34 & 754 & 62.93 & 1035 & 64.63 & 740 \\
OpenAI-o3-mini & 66.57 & 280 & 63.44 & 430 & 61.44 & 596 & 63.82 & 435 \\
OpenAI-o1 & 67.14 & 302 & 64.26 & 432 & 62.49 & 578 & 64.63 & 437 \\
QwQ-32B & 62.91 & 577 & 61.92 & 758 & 60.52 & 844 & 61.78 & 726 \\
General-CoT & 61.54 & 311 & 60.27 & 433 & 56.53 & 551 & 59.45 & 432 \\
\textbf{\texttt{TwT-Qwen2.5-7B-Instruct}} & 67.93 & 216 & 64.85 & \textbf{272} & 62.78 & \textbf{332} & 65.19 & \textbf{273} \\
\textbf{\texttt{TwT-Qwen2.5-14B-Instruct}} & \textbf{67.98} & \textbf{207} & \textbf{65.23} & 300 & \textbf{63.26} & 378 & \textbf{65.49} & 295 \\
\bottomrule
\end{tabular}%
}
\caption{In-domain performance by difficulty level.}
\label{tab:difficulty}
\end{table}

To better understand how models adapt their reasoning behavior to translation difficulty, we group the in-domain test set into three difficulty levels (Easy, Medium, Hard) estimated by DeepSeek-V3 (prompt in Figure~\ref{fig:difficulty_eval}).
Table~\ref{tab:difficulty} reports the average quality and the response length for each group, averaged across all domains.
Results indicate that \textbf{\texttt{TwT}} effectively addresses the reasoning redundancy of General-CoT through resource-rational allocation. On Easy inputs, it reduces token usage by 33\% while improving quality by $+$6.4 points; conversely, on Hard inputs, it focuses on performance, achieving a substantial $+$6.7 points quality gain.
Moreover, \textbf{\texttt{TwT}} models consistently outperform all SOTA LRMs across all difficulty levels in terms of quality, while maintaining significantly shorter reasoning traces—reducing average token usage by 32\% compared to OpenAI-o3-mini and by 60\% compared to Gemini-2.0-Flash-Thinking.
This highlights \textbf{\texttt{TwT}}'s ability to generate concise, difficulty-aware reasoning while reducing overthinking.

\subsection{Reasoning Collapse Mitigation}
\label{sec:diversity_analysis}
\begin{figure}[t]
    \centering
    \includegraphics[width=\linewidth]{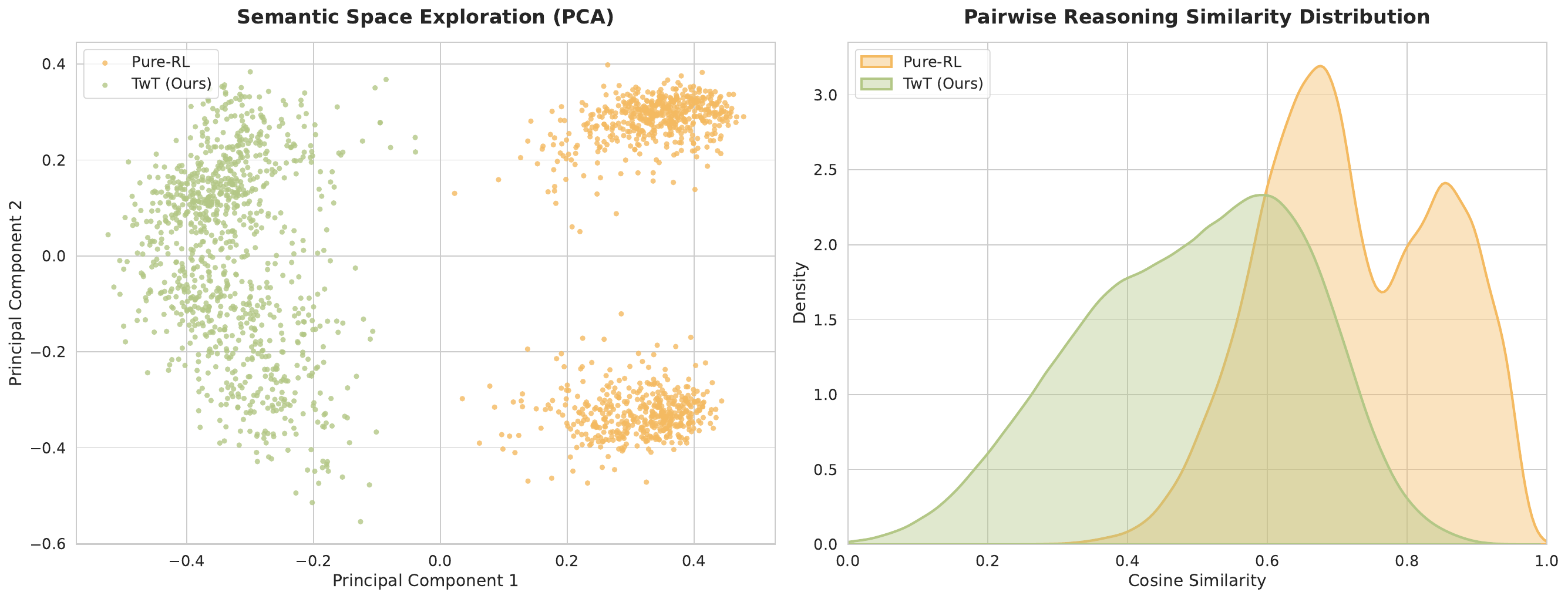}
    \caption{CoT trace similarity comparison. Pure RL vs \textbf{\texttt{TwT}} in space (Left) and distribution (Right).}
    \label{fig:diversity}
\end{figure}

In Section~\ref{sec:pure_rl_challenge}, we identified a critical failure mode of pure RL, where reasoning rapidly degenerates into shallow, repetitive templates. To verify whether \textbf{\texttt{TwT}} successfully mitigates this reasoning collapse, we conduct a semantic diversity analysis across 15 domains. We compute pairwise cosine similarities of generated reasoning traces using multilingual Sentence-BERT\footnote{sentence-transformers/paraphrase-multilingual-MiniLM-L12-v2}. 
Pure RL exhibits severe redundancy with a mean similarity of 0.89, whereas \textbf{\texttt{TwT}} significantly reduces this metric to 0.51.
This divergence is visually corroborated by Figure~\ref{fig:diversity}, where the PCA projection (Left) shows pure RL confined to tight, isolated clusters compared to the broad semantic manifold of \textbf{\texttt{TwT}}, and the similarity histogram (Right) confirms that \textbf{\texttt{TwT}} diffuses the sharp redundancy peak of the baseline into a balanced distribution. 
These results show that \textbf{\texttt{TwT}} overcomes template dependency and encourages genuine reasoning.

\subsection{Further Analysis}
Appendix~\ref{apd:further_analysis} provides additional analyses of \textbf{\texttt{TwT}}, including MQM error types, KL ablation, training dynamics, domain-aware prompting, inference cost, language consistency, and failure cases.

\section{Conclusion}

In this work, we present \textbf{\texttt{TwT}}, a resource-rational translation model that adapts reasoning effort to input difficulty. \textbf{\texttt{TwT}} combines difficulty-aware SFT and hybrid-reward RL to balance System~1 and System~2 behavior. Evaluated across diverse domains and languages, \textbf{\texttt{TwT}} matches or surpasses SOTA LRMs while reducing token usage by 32–60\%, validating the effectiveness of aligning translation with human reasoning economy.

\section*{Limitations}

While \textbf{\texttt{TwT}} achieves robust performance across multiple domains, several limitations remain. 
First, the RL training data is randomly sampled without controlling for difficulty distribution, which may result in an imbalanced mix of easy, medium, and hard inputs.
Second, the reasoning traces distilled from proprietary LLMs (e.g., DeepSeek-R1) may carry over implicit biases or domain preferences inherent in those models. Although our current setup yields consistent improvements, such biases could influence the reasoning behavior or stylistic tendencies of \textbf{\texttt{TwT}}.
Finally, our current reward design does not incorporate difficulty-aware reward shaping.
In particular, no length-based reward is applied to encourage concise reasoning on simple inputs and more detailed analysis for complex ones.
Incorporating such adaptive rewards may further enhance the model's ability to adjust reasoning depth based on input complexity in MDMT.
We leave this direction for future work.

\section*{Acknowledgment}
This work is supported by the National Science and Technology Major Project (Grant No. 2022ZD0116101),
the National Natural Science Foundation of China (NSFC) under Grant No. 62206295, 
the Major Scientific Research Project of the State Language Commission in the 13th Five-Year Plan (Grant No. WT135-38), the public technology service platform project of Xiamen City (No. 3502Z20231043). In addition, we used a large language model to assist in polishing the visualizations in Figure~\ref{fig:sys1_sys2_case} and generating certain decorative visual elements in Figure~\ref{fig:method}.

\bibliography{custom,anthology}

\appendix

\section{GRPO Algorithm}
\label{apd:grpo}
GRPO~\cite{shao2024deepseekmath} extends PPO~\citep{schulman2017ppo} by removing the dependency on a value model and instead leveraging group-wise relative rewards estimation among sampled responses for more stable and efficient policy updates.
Given a query $\mathbf{x}$, the model samples a group of $G$ responses $\{\mathbf{y}_i\}_{i=1}^G$, each scored with reward $r_i$. The normalized advantage for each sample is computed as:
\begin{equation}
A_i = \frac{r_i - \text{mean}(\{r\}_{j=1}^G)}{\text{std}(\{r\}_{j=1}^G)}.
\label{eq:adv}
\end{equation}
Then GRPO optimizes the policy model $\pi_{\theta}$ by maximizing the following objective:
\begin{equation}
\begin{aligned}
\mathcal{J}_{\text{GRPO}}(\theta) 
&= \mathbb{E}_{\mathbf{x} \sim \mathcal{D},\, \{\mathbf{y}_i\}_{i=1}^G \sim \pi_{\theta_{\text{old}}}(\cdot \mid \mathbf{x})} \\
&\Biggl(\frac{1}{G} \sum_{i=1}^G \min \Bigl(\frac{\pi_{\theta}(\mathbf{y}_i \mid \mathbf{x})}{\pi_{\theta_{\text{old}}}(\mathbf{y}_i \mid \mathbf{x})}\,A_i,\, \\
&\text{clip} \Bigl(\frac{\pi_{\theta}(\mathbf{y}_i \mid \mathbf{x})}{\pi_{\theta_{\text{old}}}(\mathbf{y}_i \mid \mathbf{x})}, 1-\epsilon, 1+\epsilon
    \Bigr)
    A_i
  \Bigr)  \\
  &- \beta D_{\text{KL}}\bigl(\pi_{\theta}\,\big\|\,\pi_{\text{ref}}\bigr)
\Biggr),
\end{aligned}
\label{eq:grpo}
\end{equation}
where $\pi_{\theta_{\text{old}}}$ and $\pi_\theta$ are the old and current policies, $\epsilon$ is the PPO clipping threshold, and $\beta$ controls the weight of the KL regularization.

\begin{figure}[t]
\centering
\begin{tcolorbox}[title=Template, label=prompt_template]
A conversation between User and Assistant. The user asks a translation question, and the Assistant solves it. The Assistant first thinks about the translation reasoning process in the mind, and then provides the final translation. The translation reasoning process and the final translation are enclosed within <think> </think> and <answer> </answer> tags, respectively, i.e., <think> translation reasoning process here </think> <answer> final translation here </answer>.\textbackslash n\textbackslash n User: \{Translation question\}.\textbackslash n Assistant: <think>
\end{tcolorbox}
\caption{Template for pure RL in MT task.}
\label{fig:prompt_zero}
\end{figure}

\begin{figure*}[t]
\centering
\includegraphics[width=0.8\textwidth]{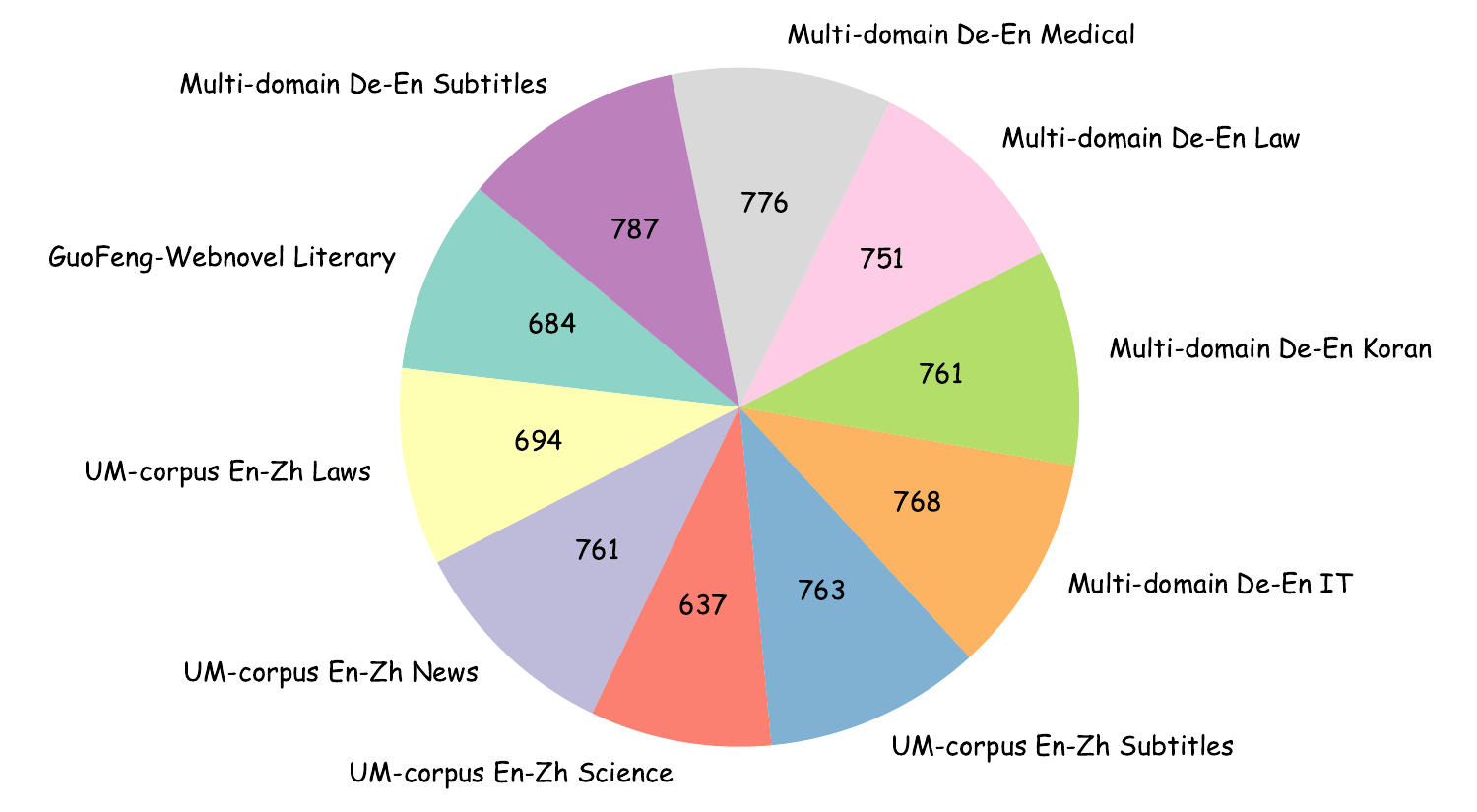}
\caption{Distribution of our curated difficulty-adaptive Long CoT data.}
\label{fig:cot_data_dist}
\end{figure*}

\section{Datasets}
\label{apd:dataset}
\subsection{Details of Cold Start Data}
For cold-start SFT, we collect a curated dataset of about 7K difficulty-adaptive Long CoT examples spanning 10 domains and three major translation directions: De$\rightarrow$En, En$\rightarrow$Zh, and Zh$\rightarrow$En. The data is constructed via domain-aware generation with DeepSeek-R1 followed by difficulty-adaptive rewriting with GPT-4o (see Section~\ref{sec:cold_start}). Detailed dataset statistics are presented in Figure~\ref{fig:cot_data_dist}.

\subsection{Details of RL Training Data}
\label{apd:rl_data}

We collect a diverse MDMT dataset for RL training across languages and domains. Specifically, we sample from the following sources:
\begin{itemize}
\item The German-English multi-domain dataset~\citep{aharoni-goldberg-2020-unsupervised}, including five distinct domains: IT, Law, Medical, Koran, and Subtitles.

\item The English-Chinese UM-Corpus~\citep{tian-etal-2014-um}, covering four domains: News, Laws, Subtitles, and Science.

\item The Chinese-English GuoFeng-Webnovel dataset~\citep{wang-etal-2023-findings,wang-etal-2024-findings} from WMT23 and WMT24 literary translation tasks, representing the Literary domain.
\end{itemize}
For each domain, we randomly select 2K sentence pairs with a minimum source sentence length of 20 words (characters for Chinese) to ensure meaningful reasoning potential.
This results in a total of 20K training samples used for RL training.

\subsection{In-Domain Test Data}
For in-domain evaluation, we use the official test sets associated with the corpora in Appendix~\ref{apd:rl_data}.
For the Literary domain, we merge the \texttt{valid\_1}, \texttt{valid\_2}, \texttt{test\_1}, and \texttt{test\_2} subsets to form a comprehensive test set.
The data statistics is illustrated in Table~\ref{tab:in-domain_test_data}.

\begin{table}[t]
\centering
\resizebox{\linewidth}{!}{%
\begin{tabular}{lc lc}
\toprule
\textbf{Domain} & \textbf{Num} & \textbf{Domain} & \textbf{Num} \\
\midrule
En$\rightarrow$Zh Laws & 456 & De$\rightarrow$En IT & 2000 \\
En$\rightarrow$Zh Subtitles & 597 & De$\rightarrow$En Koran & 2000 \\ 
En$\rightarrow$Zh Science & 503 & De$\rightarrow$En Medical & 2000 \\
En$\rightarrow$Zh News & 1500 & De$\rightarrow$En Law & 2000 \\
Zh$\rightarrow$En Literary & 3038 & De$\rightarrow$En Subtitles & 2000 \\
\bottomrule
\end{tabular}
}
\caption{In-domain test sets and the number of samples for En$\leftrightarrow$Zh and De$\rightarrow$En translation tasks.}
\label{tab:in-domain_test_data}
\end{table}

\subsection{Out-of-Domain Test Data}

For out-of-domain evaluation, we consider a diverse set of test sets spanning multiple language pairs and domains.
Specifically, the Conversation, Ecommerce, and Social domains are drawn from the WMT22 shared tasks~\cite{kocmi-etal-2022-findings}.
The Culture domain is sourced from the CAMT dataset~\cite{yao-etal-2024-benchmarking}, and the CommonSense domain comes from the CommonMT benchmark~\cite{he-etal-2020-box}.
The data statistics is illustrated in Table~\ref{tab:out-of_domain_test_data}.

\begin{table}[t]
\centering
\resizebox{\linewidth}{!}{%
\begin{tabular}{lc lc}
\toprule
\textbf{Domain} & \textbf{Num} & \textbf{Domain} & \textbf{Num} \\
\midrule
En$\rightarrow$Zh Conversation & 484 & En$\rightarrow$Zh Social & 511 \\
En$\rightarrow$Zh Ecommerce & 530 & En$\rightarrow$Zh Culture & 778 \\
Zh$\rightarrow$En CommonSense & 1200 & Zh$\rightarrow$En Conversation & 349 \\
Zh$\rightarrow$En Social & 491 & Zh$\rightarrow$En Ecommerce & 518 \\
De$\rightarrow$En Conversation & 462 & De$\rightarrow$En Social & 515 \\
De$\rightarrow$En Ecommerce & 501 & & \\
\bottomrule
\end{tabular}
}
\caption{Out-of-domain test sets and sample counts for En$\leftrightarrow$Zh and De$\rightarrow$En translation tasks.}
\label{tab:out-of_domain_test_data}
\end{table}

\begin{table*}[t]
  \centering
  \footnotesize

  \begin{tabular}{lccccc}
    \toprule
    \textbf{Model} & \textbf{Accuracy ($\downarrow$)} & \textbf{Style ($\downarrow$)} & \textbf{Fluency ($\downarrow$)} & \textbf{Terminology ($\downarrow$)} & \textbf{Non-translation ($\downarrow$)} \\
    \midrule
    DeepSeek-V3                 & 50.55 & 30.50 & 6.95 & 11.90 & 0.09 \\
    Gemini-2.0-Flash            & 52.55 & 29.55 & 6.91 & \textbf{10.81} & 0.18 \\
    GPT-4o                      & 50.52 & 30.76 & 7.11 & 11.40 & 0.21 \\
    DeepSeek-R1                 & 50.13 & 30.97 & 6.71 & 12.10 & 0.09 \\
    Gemini-2.0-Flash-Thinking   & 52.31 & 29.70 & \textbf{6.54} & 11.31 & 0.13 \\
    OpenAI-o3-mini              & \textbf{49.82} & 30.70 & 7.13 & 12.23 & 0.11 \\
    OpenAI-o1                   & 50.68 & 29.69 & 7.12 & 12.37 & 0.14 \\
    \textbf{\texttt{TwT-Qwen2.5-14B-Instruct}}     & 55.02 & \textbf{26.37} & 6.89 & 11.64 & \textbf{0.08} \\
    \bottomrule
  \end{tabular}
  \caption{MQM-based error analysis across different LLMs and LRMs.}
  \label{tab:error_analysis}
\end{table*}

\subsection{Domain Diversity Design}
To ensure comprehensive evaluation and robust generalization of \textbf{\texttt{TwT}} across diverse translation scenarios, we designed the dataset to reflect a broad range of linguistic and contextual complexity, including: high-resource domains (e.g., News); low-resource scenarios (e.g., Koranic texts); terminology-heavy fields (e.g., IT, Law, Medicine, Science, E-commerce); context-sensitive domains (e.g., Culture, CommonSense); stylistically demanding content (e.g., Literature); noisier or informal genres (e.g., Subtitles, Conversation, Social media).
This diversity ensures comprehensive evaluation across domain specificity, reasoning difficulty, and stylistic variation. 

\subsection{Multilingual Test Data}
\label{apd:multilingual}

\begin{table}[t]
\centering
\resizebox{\linewidth}{!}{%
\begin{tabular}{ll ll ll}
\toprule
ode & Language & Code & Language & Code & Language \\
\midrule
afr & Afrikaans & als & Albanian & amh & Amharic \\
asm & Assamese & bel & Belarusian & ben & Bengali \\
bos & Bosnian & bul & Bulgarian & cat & Catalan \\
cym & Welsh & ekk & Estonian & ell & Greek \\
epo & Esperanto & eus & Basque & fil & Filipino \\
gle & Irish & glg & Galician & guj & Gujarati \\
hau & Hausa & heb & Hebrew & hrv & Croatian \\
hye & Armenian & ind & Indonesian & jav & Javanese \\
kan & Kannada & kat & Georgian & kaz & Kazakh \\
khk & Mongolian & khm & Khmer & kir & Kyrgyz \\
lao & Lao & lit & Lithuanian & lvs & Latvian \\
mal & Malayalam & mar & Marathi & mkd & Macedonian \\
mya & Burmese & npi & Nepali & pan & Punjabi \\
pbt & Pashto & plt & Malagasy & san & Sanskrit \\
sin & Sinhala & slk & Slovak & slv & Slovenian \\
som & Somali & srp & Serbian & sun & Sundanese \\
swh & Swahili & tam & Tamil & tel & Telugu \\
tha & Thai & tur & Turkish & uig & Uyghur \\
urd & Urdu & vie & Vietnamese & xho & Xhosa \\
ydd & Yiddish & zsm & Malay &  &  \\
\bottomrule
\end{tabular}
}
\caption{The 59 unseen languages $\mathcal{L}_{\mathrm{unseen}}$ used for En$\leftrightarrow$X evaluation after filtering FLORES+ by (i) post-training language coverage of evaluated backbones and (ii) COMET/CometKiwi language support.}
\label{tab:unseen_lang_list}
\end{table}

For unseen-language evaluation, we adopt the FLORES+ benchmark~\citep{nllb-24} and construct an \emph{unseen} language set to minimize evaluation leakage from the languages already involved in our baseline training data (Table~\ref{tab:result_lang}). Concretely, we first exclude all languages that appear in the baseline training coverage, including Chinese (zh), English (en), German (de), French (fr), Spanish (es), Portuguese (pt), Italian (it), Russian (ru), Korean (ko), Dutch (nl), Czech (cs), Icelandic (is), Ukrainian (uk), Hindi (hi), Japanese (ja), Polish (pl), Swedish (sv), Hungarian (hu), Romanian (ro), Danish (da), Norwegian (no), and Finnish (fi). This step ensures that the En$\leftrightarrow$X results reflect generalization to genuinely unseen languages rather than memorization of language-specific post-training signals.
Next, to enable consistent computation of COMET and CometKiwi across all unseen directions, we further restrict the remaining FLORES+ languages to those supported by our COMET/CometKiwi scorers. After these two filters, we obtain a final unseen-language set consisting of 59 languages (Table~\ref{tab:unseen_lang_list}).

The \textit{seen} languages—German (de), English (en), and Chinese (zh)—are used across all baseline training sets, and are covered by the following datasets: the German-English multi-domain dataset~\citep{aharoni-goldberg-2020-unsupervised}, the English-Chinese UM-Corpus~\citep{tian-etal-2014-um}, and the Chinese-English GuoFeng-Webnovel dataset.

\section{Training Data Scale of MT Baselines}
\label{apd:mt_data_scale}
To ensure a fair comparison, we report the training data size used by each MT-specialized baseline. While \textbf{\texttt{TwT}} and most reasoning-augmented models, including MT-R1-Zero-7B, CoT-FT-7B, SFT-Parallel, and mExTrans-7B, are trained on approximately 27K examples, several other models leverage significantly larger corpora. For instance, TowerInstruct series is trained on 637K examples, Tower-Plus-9B on 286K, and ALMA-R on 21K. SSR-X-Zero-7B is trained on a notably smaller subset of 13K instances.

\section{API Details}
\label{appendix:sota_api}

The following APIs were used to access the SOTA LLMs and LRMs evaluated in our experiments:

\begin{itemize}
    \item \textbf{OpenAI:} gpt-4o-2024-11-20, o1-2024-12-17, o3-mini-2025-01-31, and gpt-5-2025-08-07
    \item \textbf{DeepSeek:} deepseek-chat-2024-12-26, and deepseek-reasoner
    \item \textbf{Gemini:} gemini-2.0-flash, and gemini-2.0-flash-thinking-exp-01-21
\end{itemize}

\section{Experimental Details for Preliminary Analysis}
\label{apd:preliminary}

\subsection{Setup for Pure RL}
\label{appendix:r1-zero}
\noindent \textbf{Training Data.}
We construct a diverse MDMT training data with 20K samples by randomly sampling 2K sentence pairs from 10 domains, spanning three major language pairs: De$\rightarrow$En~\citep{aharoni-goldberg-2020-unsupervised}, En$\rightarrow$Zh~\citep{tian-etal-2014-um}, and Zh$\rightarrow$En~\citep{wang-etal-2023-findings,wang-etal-2024-findings}.
We select only examples with source sentences longer than 20 words (characters for Chinese) to encourage meaningful reasoning.
Evaluation is conducted using the test sets provided by the original datasets.

\noindent \textbf{Training Template.} 
To guide the base model toward producing translation-relevant reasoning behavior, we modify the template used in DeepSeek-R1-Zero~\citep{deepseekai2025deepseekr1} to better fit the translation task. The chat template is shown in Figure~\ref{fig:prompt_zero}.

\noindent \textbf{Reward Design.} 
We follow DeepSeek-R1~\citep{deepseekai2025deepseekr1} in using two reward types.

\noindent \textit{Format Reward:} 
We apply regex matching to check whether the model responses adhere to the specified format. 
The format reward $r_f$ is set to 1 if the format is correct, and -1 otherwise.

\noindent \textit{Quality Reward:} 
We experiment with several quality metrics to evaluate the final translation output.
Specifically, we consider BLEU\footnote{\url{https://github.com/mjpost/sacrebleu}}~\citep{papineni-etal-2002-bleu}, a metric based on n-gram lexical overlap, and COMET\footnote{\texttt{Unbabel/wmt22-comet-da}}~\citep{rei-etal-2020-comet} and CometKiwi\footnote{\texttt{Unbabel/wmt22-cometkiwi-da}}~\citep{rei-etal-2022-cometkiwi}, neural metrics that estimate semantic similarity using contextual embeddings through reference-based or reference-free scoring.
All metric scores are normalized to the range $[0,1]$ before being used in the reward function.
During training, the quality reward $r_{q}$ is computed based on one or more of these metrics.
It is only applied when the output format is correct;
otherwise, we assign a fixed penalty of $r_{q}=-2$.
The final reward used during RL training is the sum of the format and quality components: $r=r_{f}+r_{q}$.

\noindent \textbf{RL Optimization.} 
We train the model using GRPO algorithm~\citep{shao2024deepseekmath}, which improves stability over PPO~\citep{schulman2017ppo} by leveraging group-wise relative advantages rather than explicit value functions.
See Appendix~\ref{apd:grpo} for full formulation.

\noindent \textbf{Model Training.}
We conduct experiments using Qwen2.5-7B-Instruct~\citep{qwen2025qwen25} as the base model, and train it for 1 epoch on our collected 20k MDMT data. To investigate the effectiveness of different reward signals, we experiment with five quality reward variants: BLEU, COMET, CometKiwi, BLEU+COMET, and BLEU+CometKiwi.

\subsection{Setup for RL with SFT}
\label{appendix:r1-cold-start}
\noindent \textbf{Training Data.}
We distill reasoning traces from the DeepSeek-R1 model~\citep{deepseekai2025deepseekr1} and apply additional quality filtering, resulting in about 7K high-quality Long CoT examples for MDMT.

\noindent \textbf{Model Training.}
We fine-tune Qwen2.5-7B-Instruct~\citep{qwen2025qwen25} on the constructed Long CoT dataset for 1 epoch, followed by RL training on the 20K multi-domain data for 1 epoch, using a composite reward of BLEU and COMET. General-CoT and Domain-CoT share the same training pipeline and hyperparameters, differing only in the prompt used for SFT data construction: General-CoT uses "\texttt{Translate the following \{src\_lang\} text into \{tgt\_lang\}.}", while Domain-CoT uses "\texttt{Translate the following \{src\_lang\} text into \{tgt\_lang\} while maintaining the domain style of the source text.}"

\section{Further Analysis}
\label{apd:further_analysis}

\subsection{Error-Type Analysis under the MQM Framework}
\label{apd:error_type}
To further investigate what types of translation errors reasoning helps address, we conducted a detailed analysis based on the MQM (Multidimensional Quality Metrics) framework. 
For error classification, we used an external LLM as an annotator, excluding "Source Error" as it pertains to low-quality source data.
Given the distinct challenges of MDMT compared to general MT, we extended error types such as "Style Inconsistency", "Cross-domain Confusion", and "Terminology Misuse" to better capture the advantages of reasoning in MDMT. 
As shown in Table~\ref{tab:error_analysis}, reasoning-enhanced LRMs lead to lower error rates in Style, Fluency, Terminology, and Non-translation Errors. We believe this is due to the structured reasoning process, which contributes to a more human-like translation process and better error reflection, in contrast to the pattern-matching of traditional bilingual fine-tuning. Notably, after incorporating domain-aware CoT data, \textbf{\texttt{TwT}} reduced style errors by 3.18\%, significantly improving the model's style adaptation. This highlights the effectiveness and applicability of our approach in MDMT.

\begin{table}[t]
\centering
\resizebox{\linewidth}{!}{%
\begin{tabular}{lccccc}
\toprule
$\beta$ & BLEU & COMET & KIWI & Quality & Lens. \\
\midrule
0 & 30.07 & 82.35 & 81.20 & 64.54 & 130 \\
0.001 & 30.58 & 82.52 & 81.21 & 64.77 & 278 \\
0.005 & 30.79 & 82.48 & 81.28 & 64.85 & 273 \\
0.01 & 30.37 & 82.49 & 81.40 & 64.75 & 282 \\
0.02 & 30.24 & 82.37 & 81.29 & 64.63 & 254 \\
0.05 & 29.71 & 82.23 & 81.24 & 64.39 & 275 \\
\bottomrule
\end{tabular}%
}
\caption{In-domain results with different KL coefficient.}
\label{tab:kl_weight}
\end{table}

\subsection{KL Coefficient Analysis.}
\label{apd:kl_ablation}
Recent study~\cite{yu2025dapoopensourcellmreinforcement} suggests that removing KL regularization can enhance Long CoT reasoning.
To examine whether this holds in the MDMT setting, we investigate the effect of the KL coefficient $\beta$ in Table~\ref{tab:kl_weight}.
However, our findings diverge from this conclusion: setting $\beta$ to 0—completely removing the KL constraint—results in significantly shorter and less informative outputs. 
In contrast, small non-zero values (\emph{e.g.}, $\beta=0.001$ or $0.005$) achieve the best overall quality scores while maintaining reasonable response lengths. 
Larger values (\emph{e.g.}, $\beta=0.02$ or $0.05$) overly constrain the policy, slightly degrading translation quality and increasing length instability.
 These results reflect the importance of balancing generation stability and flexibility: moderate KL regularization helps suppress degenerate behavior while preserving adaptive, difficulty-aware reasoning traces.

\subsection{Analysis of Training Dynamics.}
\label{apd:training_dynamics}
To better understand how RL shapes \textbf{\texttt{TwT}}'s reasoning and translation behavior, we analyze the training dynamics of \textbf{\texttt{TwT-7B}} and \textbf{\texttt{TwT-14B}} from two complementary perspectives.

\begin{figure}[t]
    \centering
    \includegraphics[width=\linewidth]{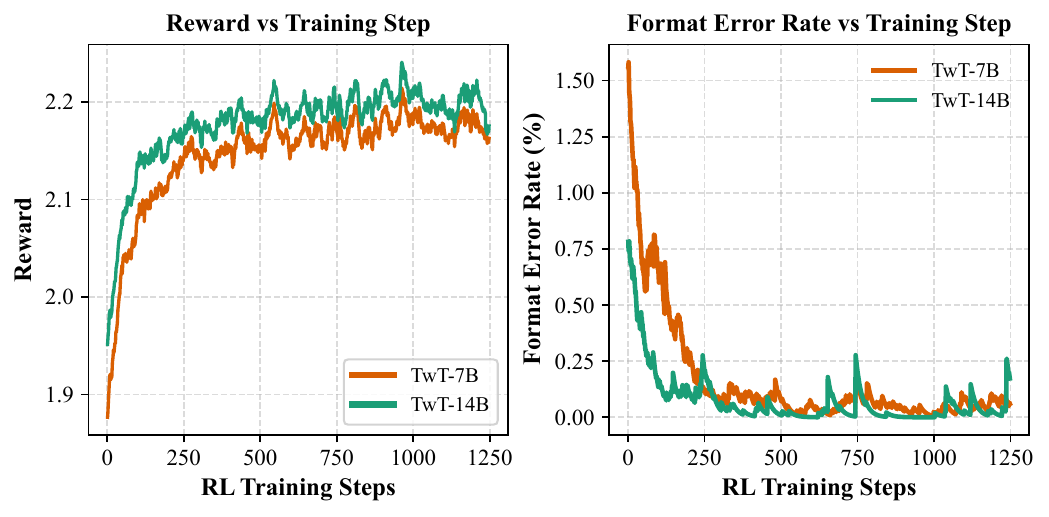}
    \caption{Average reward and format error rate over RL training steps.}
    \label{fig:reward_curves}
\end{figure}

\begin{figure*}[t]
\centering
\subfigure[Training dynamics for \textbf{\texttt{TwT-7B}} and \textbf{\texttt{TwT-14B}}.]{
\label{fig:traing_dynamics}
\includegraphics[width=\textwidth]{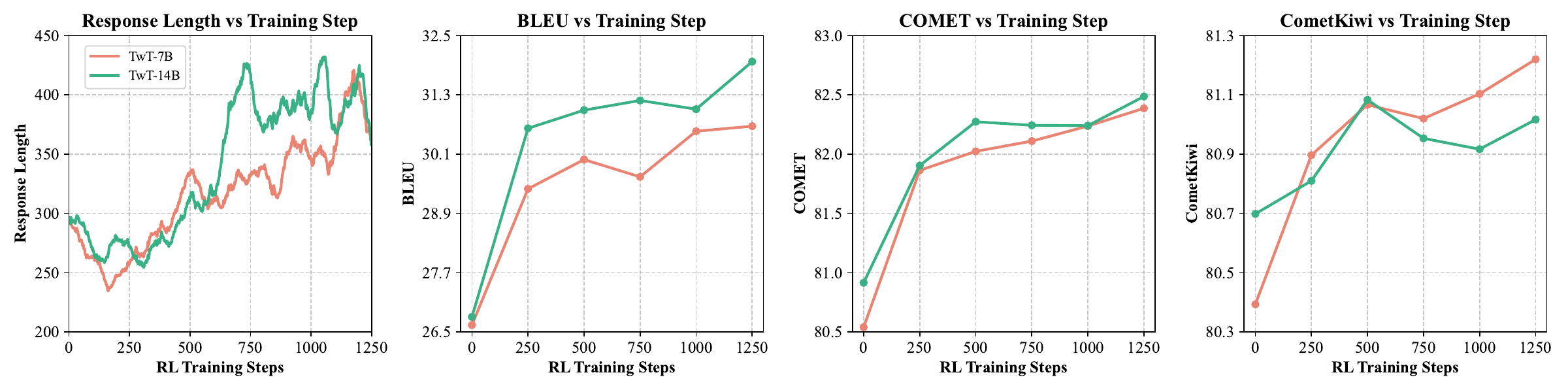}
}
~
\subfigure[Training dynamics for \textbf{\texttt{TwT-7B}} under different difficulty level.]{
\label{fig:curves_7B}
\includegraphics[width=\textwidth]{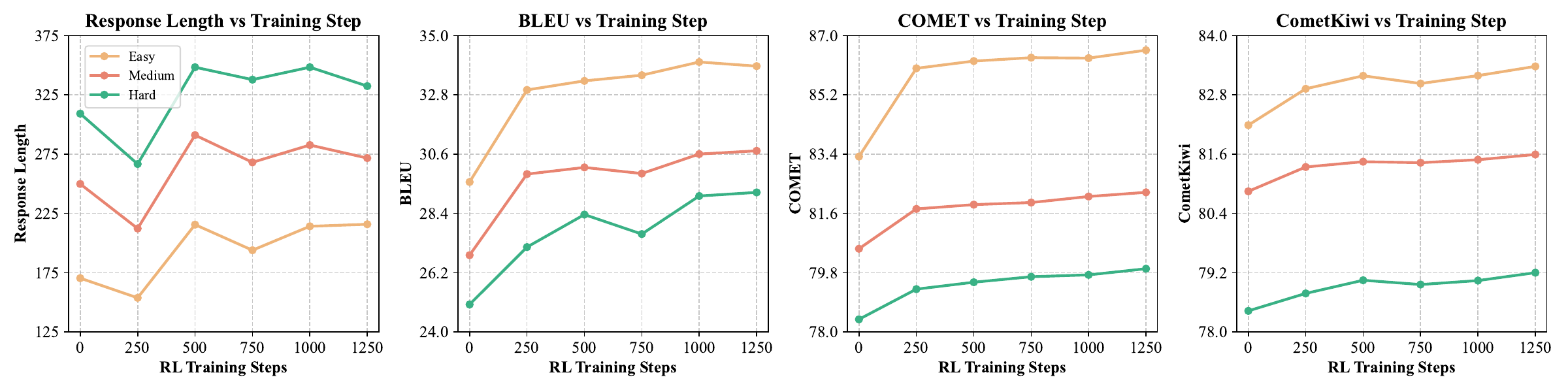}
}
~
\subfigure[Training dynamics for \textbf{\texttt{TwT-14B}} under different difficulty level.]{
\label{fig:curves_14B}
\includegraphics[width=\textwidth]{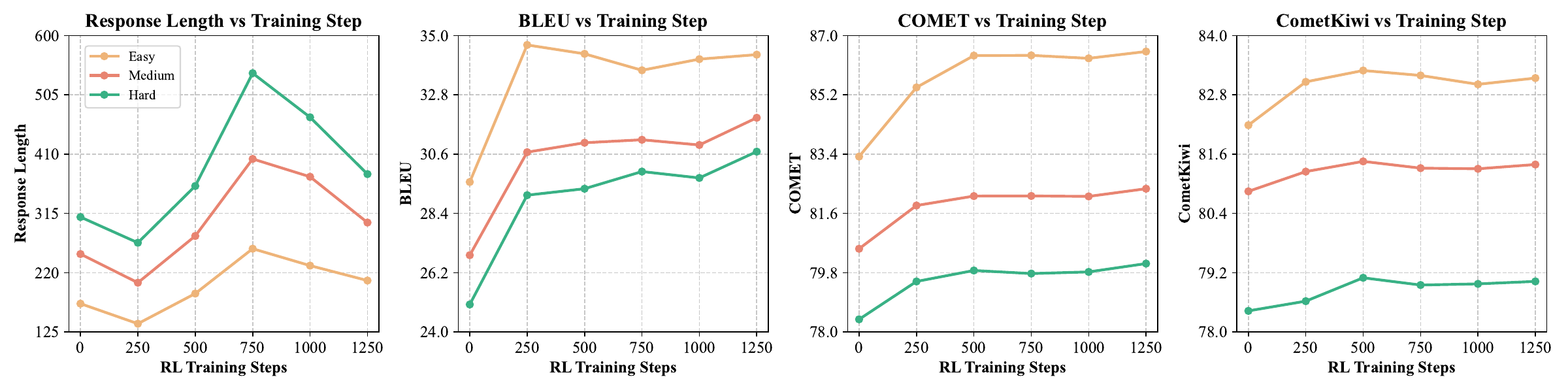}
}
\caption{Training dynamics for \textbf{\texttt{TwT}} models.}
\label{fig:training_curves}
\end{figure*}

\paragraph{Analysis of Reward and Format Stability}
Figure~\ref{fig:reward_curves} further illustrates the reinforcement learning dynamics of \textbf{\texttt{TwT-7B}} and \textbf{\texttt{TwT-14B}} in terms of average reward and format error rate. 
Both models show a rapid increase in reward within the first few hundred steps, followed by a stable plateau, indicating that the hybrid reward formulation enables efficient convergence toward high-quality translation reasoning. The larger \textbf{\texttt{TwT-14B}} model consistently achieves higher rewards throughout training, suggesting stronger optimization capacity and better utilization of the reward signal.
In contrast, the format error rate decreases sharply during the initial phase and remains near zero thereafter, demonstrating that the format reward effectively reduces structural inconsistencies and output anomalies as training progresses.

\paragraph{Translation Quality and Reasoning Depth}
Figure~\ref{fig:training_curves} shows the training dynamics of \textbf{\texttt{TwT-7B}} and \textbf{\texttt{TwT-14B}} during RL. BLEU, COMET, and CometKiwi scores steadily improve, indicating that our hybrid reward effectively enhances translation quality. Interestingly, response length increases in the early and mid stages, reflecting deeper reasoning, but decreases in later steps as the model learns to maintain quality with more concise traces. This indicates convergence to efficient, difficulty-aware reasoning behavior, as shown in Figure~\ref{fig:traing_dynamics}.
Figure~\ref{fig:curves_7B} and Figure~\ref{fig:curves_14B} show the performance of \textbf{\texttt{TwT-7B}} and \textbf{\texttt{TwT-14B}} under different difficulty levels. In both models, BLEU, COMET, and CometKiwi scores are highest on easy inputs and lowest on hard ones, and response length increases with difficulty, indicating effective control of reasoning depth. Notably, \textbf{\texttt{TwT-14B}} generates shorter responses than \textbf{\texttt{TwT-7B}} on easy inputs, but longer responses on medium and hard inputs. This suggests that the larger model more effectively adapts its reasoning length to input difficulty, providing concise outputs when possible and allocating more reasoning to harder cases.

\begin{table}[t]
\centering
\caption{General prompt vs. domain-aware prompt across domains.}
\label{tab:domain-aware-prompt}
\resizebox{\linewidth}{!}{
\begin{tabular}{lccccccccc}
\toprule
\multirow{2}{*}{\textbf{Prompt}} &
\multicolumn{3}{c}{\textbf{IT}} &
\multicolumn{3}{c}{\textbf{Law}} &
\multicolumn{3}{c}{\textbf{Medical}} \\
\cmidrule(lr){2-4} \cmidrule(lr){5-7} \cmidrule(lr){8-10}
 & BLEU & COMET & KIWI
 & BLEU & COMET & KIWI
 & BLEU & COMET & KIWI \\
\midrule
General prompt & \textbf{36.66} & 83.49 & 78.68 & 39.16 & 84.97 & 82.35 & 40.23 & 83.82 & 81.81 \\
Domain-aware prompt & 36.55 & \textbf{83.57} & \textbf{79.82} & \textbf{40.57} & \textbf{85.32} & \textbf{83.54} & \textbf{41.14} & \textbf{84.06} & \textbf{82.97} \\
\bottomrule
\end{tabular}
}
\end{table}

\begin{table}[t]
\centering
\caption{SFT trained on general CoT vs. domain-aware CoT.}
\label{tab:domain-aware-cot}
\resizebox{\linewidth}{!}{
\begin{tabular}{lcccccc}
\toprule
\multirow{2}{*}{\textbf{Method}} & \multicolumn{3}{c}{\textbf{In-Domain}} & \multicolumn{3}{c}{\textbf{Out-of-Domain}} \\
\cmidrule(lr){2-4} \cmidrule(lr){5-7}
 & BLEU & COMET & KIWI & BLEU & COMET & KIWI \\
\midrule
SFT w/ general CoT      & 23.51 & 80.52 & 80.15 & 25.12 & 83.40 & 80.81 \\
SFT w/ domain-aware CoT & \textbf{23.74} & \textbf{80.74} & \textbf{80.46} & \textbf{25.67} & \textbf{83.57} & \textbf{81.18} \\
\midrule
$\Delta$ & +0.23 & +0.22 & +0.31 & +0.55 & +0.17 & +0.37 \\
\bottomrule
\end{tabular}
}
\end{table}

\subsection{Analysis of Domain-Aware Prompting}
\label{apd:domain_prompt}
To better understand the effect of domain-aware prompting, we conduct two complementary experiments that evaluate its impact from both the inference and training perspectives.

First, we investigate how prompt formulation affects translation quality when the model remains fixed (DeepSeek-R1). Two prompts are compared: a general prompt ("Translate the following {src\_lang} sentence into {tgt\_lang}.") and a domain-aware prompt ("Translate the following {src\_lang} text into {tgt\_lang} while maintaining the domain style of the source text."). 
As shown in Table~\ref{tab:domain-aware-prompt}, the comparison across three representative domains (IT, Law, Medical) demonstrates that domain-aware prompt explicitly instructs the model to infer and preserve domain-specific stylistic and terminological features. This better activates the reasoning capacity and leads to more accurate domain-aware translation. These findings motivated us to adopt domain-aware prompts throughout \textbf{\texttt{TwT}}'s training data curation phase, allowing domain signals to be explicitly injected during learning.

Second, we evaluate the effect of using different prompts to construct Long-CoT data for SFT. Specifically, one dataset is generated using the general prompt, while another uses the domain-aware prompt to elicit domain-specific reasoning traces.
As shown in Table~\ref{tab:domain-aware-cot}, domain-aware CoT data lead to consistent improvements across all automatic evaluation metrics in both in-domain and out-of-domain scenarios. These results highlight that incorporating domain context enhances the model's reasoning robustness and cross-domain generalization.

\begin{figure}[t]
    \centering
    \includegraphics[width=\linewidth]{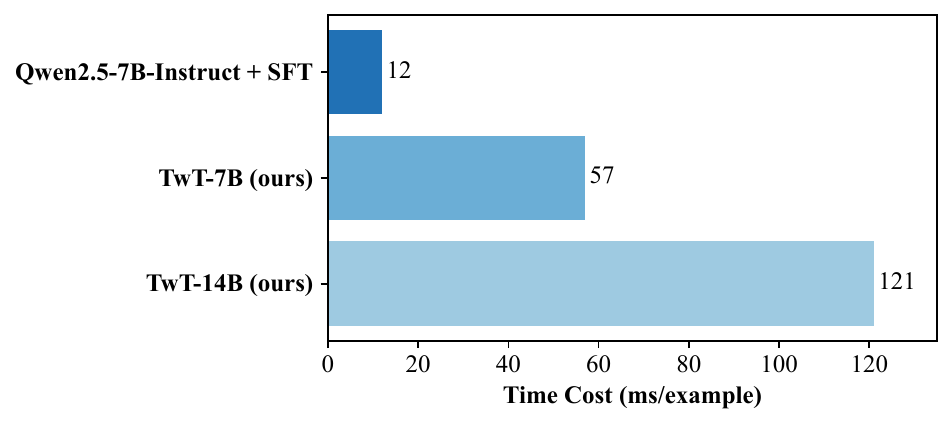}
    \caption{Inference time comparison. }
    \label{fig:inference_time}
\end{figure}

\subsection{Computational Cost Analysis}
\label{apd:cost_analysis}
To assess the computational efficiency of our reasoning-based translation paradigm, we compare the average inference time per example across different models.
As shown in Figure~\ref{fig:inference_time}, the reasoning-augmented \textbf{\texttt{TwT}} models incur additional computational overhead compared with the non-reasoning baseline (Qwen2.5-7B-Instruct + SFT). Specifically, \textbf{\texttt{TwT-7B}} and \textbf{\texttt{TwT-14B}} require 57 ms and 121 ms per example, respectively, compared to 12 ms for the baseline.
Despite the higher inference cost, the time overhead remains acceptable given the significant improvements in translation quality, reasoning accuracy, and style adaptation demonstrated in previous sections. 

\subsection{Language Consistency Discussion}
\label{apd:lang_consist}
To verify that \textbf{\texttt{TwT}}'s reasoning traces are linguistically coherent and aligned with the target translation language, we conducted a systematic analysis of language consistency throughout the data curation and training stages.
We employed a language identification tool (fastText) to detect potential cross-lingual inconsistencies in reasoning traces, such as mixing English reasoning with Chinese or German summaries.
In the initial R1-distilled CoT data, 9.26\% (684/7382) of the reasoning traces contained incorrect languages. After GPT-4o rewriting, the inconsistency dropped sharply to 3.78\% (279/7382).
After SFT, the inconsistency rate on the test set further decreased to 0.04\% (7/16094), and it reached 0\% after the RL stage.

\subsection{Reliability of Difficulty Estimation}
\label{apd:difficulty_reliability}

To mitigate potential bias from relying solely on GPT-4o, we validated its difficulty estimation against two other strong models (DeepSeek-V3.2, Gemini-2.5-Flash) and human experts on 100 randomly sampled instances. 
Results demonstrate robust consistency across three dimensions:
(1) \textbf{Internal Stability:} All models exhibited high self-agreement across 5 independent trials (GPT-4o: 0.92, DeepSeek: 0.93, Gemini: 0.91).
(2) \textbf{Cross-Model Agreement:} The three models reached a 90\% consensus rate after majority voting.
(3) \textbf{Human Alignment:} Crucially, GPT-4o achieved the highest correlation with professional translators (\textbf{0.86}), surpassing DeepSeek (0.77) and Gemini (0.79). 
These findings confirm that GPT-4o provides a reliable and stable proxy for human judgment in our difficulty-aware framework.

\subsection{Structured Analysis of Failure Cases}
To provide a more systematic analysis, we further examine \textbf{\texttt{TwT}}'s bad cases across all domains. Specifically, we select the 10 lowest-quality examples from each of the 15 domains (150 cases in total) and analyze their CoT trajectories. We identify two recurring error patterns.

\paragraph{Reasoning--prediction misalignment.}
In some cases, the model makes the correct translation decision in its reasoning, but the final output fails to realize it. That is, the reasoning identifies an appropriate lexical choice or phrasing, while the prediction deviates from it. This suggests a gap between reasoning and realization. A possible remedy is to introduce an additional reward penalty during GRPO to explicitly discourage divergence between the decision expressed in the \texttt{<think>} block and the output in the \texttt{<answer>} block.

\paragraph{Domain--terminology misalignment.}
Another common failure occurs when the model correctly identifies the domain but does not switch to the appropriate terminology system. In such cases, the reasoning remains at a declarative level without consistently applying domain-specific lexical choices, syntactic preferences, fixed translations, or formatting conventions. One possible remedy is to incorporate external terminology lexicons or style-template libraries to guide domain-specific realization.

\begin{table}[t]
\centering
\resizebox{\linewidth}{!}{%
\begin{tabular}{lccc}
\toprule
\textbf{Human Evaluation} & \textbf{TwT Win} & \textbf{TwT Lose} & \textbf{Tie} \\
\midrule
\textbf{\texttt{TwT-Qwen2.5-14B-Instruct}} vs. DeepSeek-R1 (671B) & 0.41 & 0.47 & 0.12 \\
\textbf{\texttt{TwT-Qwen2.5-14B-Instruct}} vs. DeepSeek-V3 (671B) & 0.275 & 0.405 & 0.32 \\
\textbf{\texttt{TwT-Qwen2.5-14B-Instruct}} vs. TowerInstruct-13B-v0.1 & \textbf{0.54} & 0.30 & 0.16 \\
\bottomrule
\end{tabular}
}
\caption{Summary of human evaluation results for TwT compared with other models.}
\label{tab:human_eval}
\end{table}

\section{Human Evaluation}
\label{apd:human_evaluation}

\subsection{Human Evaluation Analysis}
We conducted a human evaluation to complement our automatic metrics. 
Specifically, we randomly sampled 100 examples from the Zh$\leftrightarrow$En test set, selecting 20 sentences from each of five domains. 
For each instance, the source sentence and two system outputs—one from \textbf{\texttt{TwT-14B}} and one from a comparison model—were independently evaluated by three professional translators, who selected the better translation or marked a tie when the quality difference was negligible.
Table~\ref{tab:human_eval} summarizes the human evaluation results across three baselines: DeepSeek-R1, DeepSeek-V3, and TowerInstruct-13B-v0.1.
These results show that while \textbf{\texttt{TwT-14B}} slightly underperforms DeepSeek-R1 and DeepSeek-V3—expected given its much smaller size—it still achieves a strong degree of parity. Importantly, \textbf{\texttt{TwT-14B}} outperforms TowerInstruct-13B-v0.1 in more than half of the evaluated examples (54\% win rate), supporting the effectiveness of its reasoning-driven design in translation.

\subsection{Comparison with Human Translation Reasoning}
\label{apd:human_eval_case}
\begin{figure*}[!htbp]
\centering
\includegraphics[width=\textwidth]{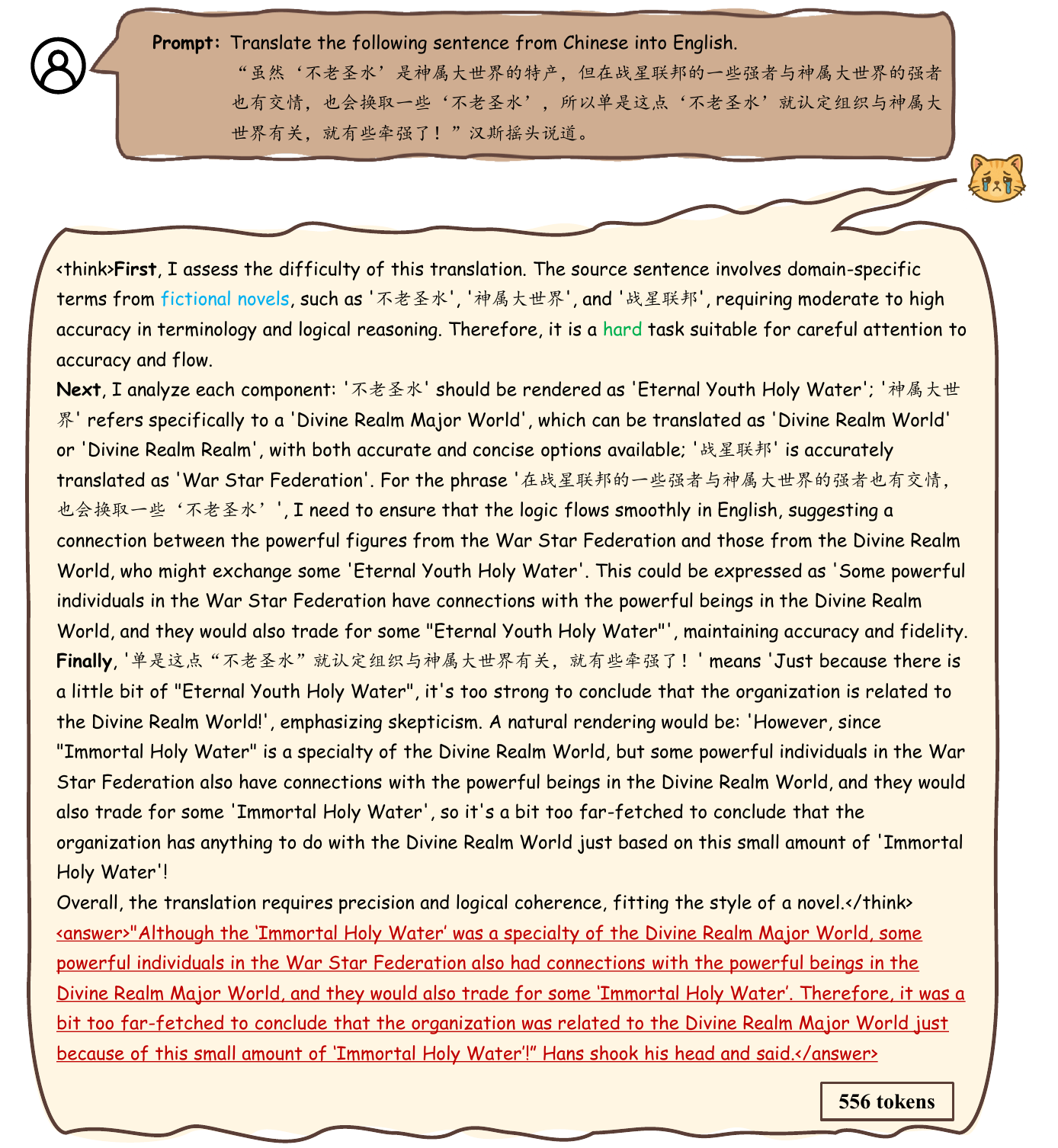}
\caption{Output examples of our \textbf{\texttt{TwT-7B}}.}
\label{fig:case_hard}
\end{figure*}

To further evaluate the alignment between \textbf{\texttt{TwT}}'s reasoning process and that of professional human translators, we conducted a case study using a complex narrative sentence from a fictional novel, as shown in Figure~\ref{fig:case_hard}.
The source sentence features multiple subordinate and concessive clauses, complex logical progression, and dense world-specific terminology—a typical example in the fictional novel domain. Such sentences require precise control of syntax, consistent terminology management, and sensitivity to logical flow and pragmatic tone.

\begin{CJK}{UTF8}{gbsn}
\paragraph{Similarities.}
\textbf{\texttt{TwT}} demonstrates several key reasoning behaviors consistent with professional translators.
\textbf{First}, it correctly identified the sentence structure and logical relations, decomposing the original into two conceptual layers: (1) background explanation of the connection between the 神属大世界 and the 战星联邦, and (2) a reasoning-based denial of the assumed link between the 组织 and 神属大世界. This decomposition mirrors the syntactic and logical analysis stage in human translation reasoning, reflecting \textbf{\texttt{TwT}}'s strong ability to detect hierarchical structure and causal relationships.
\textbf{Second}, \textbf{\texttt{TwT}} shows strong terminological consistency, accurately translating key domain-specific entities such as "Immortal Holy Water" (不老圣水), "Divine Realm Major World" (神属大世界), and "War Star Federation" (战星联邦). While minor stylistic improvements remain possible, the model's use of consistent and semantically clear terms aligns with the terminology standardization step in professional translation workflows. \textbf{\texttt{TwT}} also handled subtle pragmatic expressions such as "有交情" and "换取" with contextually appropriate equivalents ("have connections with", "trade for"), indicating an emerging awareness of functional-pragmatic equivalence—a key element in expressing communicative intent and contextual tone.
\textbf{Moreover}, before translating, \textbf{\texttt{TwT}} demonstrated an initial assessment of text difficulty and style, recognizing that the sentence belongs to a fictional novel with specialized terms. This awareness parallels a human translator's pre-translation difficulty assessment and background analysis, where genre and register are evaluated to inform strategy. \textbf{\texttt{TwT}} chose to preserve the source-style transliteration rather than domesticate the names, maintaining consistency with the fantasy universe's internal logic—an appropriate decision for this genre.
\textbf{At the structural level}, \textbf{\texttt{TwT}} effectively reconstructed the concessive–causal logic of "虽然…但…所以…" into the English pattern "Although...some...therefore...", faithfully capturing the original logical progression. This demonstrates the model's ability to rebuild syntactic and logical relations during target-language reorganization, consistent with human translators’ reasoning in structural decomposition and coherence reconstruction.
\textbf{In lexical judgment}, \textbf{\texttt{TwT}} made contextually informed choices, such as rendering "不老圣水" as "Immortal Holy Water" rather than the freer "elixir of youth" or "anti-aging water", thereby preserving the source's mythological tone. Similarly, translating "牵强" as "far-fetched" appropriately conveys the intended skepticism while maintaining stylistic naturalness. These decisions illustrate the model's ability to perform semantic disambiguation and lexical selection comparable to human translators’ third-stage reasoning process.
\textbf{Finally}, \textbf{\texttt{TwT}}'s output features a natural syntactic flow and coherent discourse structure. For instance, "some powerful individuals...had connections with..." reconstructs the information hierarchy more fluently than a literal rendering would. This indicates an emerging sense of register and stylistic adaptation, partially fulfilling the requirements of the expression and style selection stage.

\paragraph{Differences.}

\textbf{First}, \textbf{\texttt{TwT}}'s translation repeatedly uses the long-form expression such as "Divine Realm Major World", resulting in verbosity. Professional translators would typically balance terminological consistency with referential economy, adopting simplified references such as "the Divine Realm" or "that realm" to improve fluency and readability.
For instance, a revised version could read: "Although ‘Immortal Holy Water’ is a specialty of the Divine Realm, some powerful individuals in the War Star Federation have connections with that realm and can obtain it through trade." This alternative maintains precision while achieving smoother rhythm and reduced redundancy.
\textbf{Second}, \textbf{\texttt{TwT}}'s stylistic control remains limited. Although the translation is grammatically accurate, it lacks the conversational tone and personality expected in dialogue.
Since the original sentence is dialogue from Hans expressing skepticism, a professional translator would employ a more natural, idiomatic style to reflect the speaker's voice and pragmatic intent, e.g.,
"Come on, you can’t just assume they’re connected to the Divine Realm over a bit of Holy Water—that's a stretch!"
Such adaptation better reproduces conversational tone and aligns with the narrative style of fantasy dialogue.
\textbf{Third}, \textbf{\texttt{TwT}}'s cultural adaptation is surface-level. While "far-fetched" effectively captures the core meaning of "牵强," a human translator might enrich cultural accessibility by appending a light explanatory note, such as "Immortal Holy Water, a sacred elixir rumored to halt aging in the Divine Realm." This hybrid "foreignization plus domestication" strategy balances cultural retention with reader comprehension, commonly seen in professional literary translation.
\end{CJK}

\section{Prompt}
The template for pure RL (Section~\ref{sec:pure_rl_challenge}) is shown in Figure \ref{fig:prompt_zero}.
The prompt for generating difficulty-adaptive Long CoT data is provided in Figure \ref{fig:difficulty_polish_prompt}.
The prompt employed for the quantitative evaluation of reasoning efficiency is presented in Figure \ref{fig:redundancy_prompt}.
The prompt for evaluating translation difficulty is shown in Figure \ref{fig:difficulty_eval} and the detailed examples for translation difficulty evaluation are provided in Table~\ref{tab:difficulty_example}. 

\begin{figure*}[t]
\centering
\begin{tcolorbox}[colback=gray!5, colframe=black, width=\textwidth]

You are an experienced translation expert. Your task is to optimize a given initial translation thought process by simulating the thinking process of a professional human translator.
\\
\\
Human translators typically first assess the difficulty of the translation task. Based on this difficulty level, they allocate appropriate time and cognitive effort:
\begin{itemize}
\item For simple translation tasks, they provide an accurate translation quickly with concise thought process.
\item For moderate translation tasks, they reason carefully through any ambiguities to produce an accurate translation, without excessive expansion.
\item For difficult translation tasks, they conduct deep and comprehensive thinking, exploring different translation strategies, comparing and verifying options, and refining every detail to produce the best translation.
\end{itemize}
Based on this professional approach, please optimize the given translation thought process by:
\begin{itemize}
\item Integrating the assessment of translation difficulty naturally into the thought process.
\item Adjusting the depth and style of reasoning based on the difficulty level.
\item Preserving the original language style and tone of the initial translation thought process.
\end{itemize}
Finally, output the optimized thought process and the final translation in JSON format with the keys "thought" and "translation", as shown below:
\begin{verbatim}
{
  "thought": "Optimized translation thought process based on difficulty",
  "translation": "Final translation based on the corrected thought process"
}
\end{verbatim}

Source text: 

Target language: 

Initial translation thought process: 

\end{tcolorbox}
\caption{Full prompt used for generating difficulty-adaptive Long CoT data with GPT-4o.}
\label{fig:difficulty_polish_prompt}
\end{figure*}

\begin{figure*}[t]
\centering
\begin{tcolorbox}[colback=gray!5, colframe=black, width=\textwidth]

Your task is to assess the difficulty of translating a given \texttt{\{src\_lang\}} sentence into \texttt{\{tgt\_lang\}}. Please evaluate the difficulty based on the following criteria:
\begin{enumerate}
\item Sentence complexity: Determine if the sentence is a simple sentence, a compound sentence, or includes subordinate clauses and other complex structures.
\item Vocabulary difficulty: Assess whether the sentence contains commonly used words or specialized terms or slang.
\item Grammar differences: Analyze if the sentence's grammatical structure is similar to or differs significantly from \texttt{\{tgt\_lang\}}.
\item Contextual understanding: Consider whether understanding specific cultural contexts or background knowledge is necessary for accurate translation.
\end{enumerate}
The difficulty level should be rated as "easy", "medium", or "hard". Additionally, provide a brief, simple reason for the assigned difficulty level. Output the result in JSON format with the keys "level" and "reason", as follows.
\begin{verbatim}
{
  "level": "easy/medium/hard",
  "reason": "simple explanation of the difficulty level."
}
\end{verbatim}

Here is the \texttt{\{src\_lang\}} sentence: \texttt{\{src\_text\}}

\end{tcolorbox}
\caption{Full prompt used for evaluating translation difficulty with GPT-4o.}
\label{fig:difficulty_eval}
\end{figure*}

\begin{figure*}[t]
\centering
\begin{tcolorbox}[colback=gray!5, colframe=black, width=\textwidth]
\setstretch{0.9}

You are an expert model specialized in ``Chain-of-Thought (CoT) Quality Evaluation for Multi-Domain Translation Tasks.'' Your task is to compare the reasoning traces of Model A and Model B: Identify redundancy types in Model A's reasoning and determine whether Model B has successfully eliminated these redundancies.
\\
\\
Please strictly adhere to the following \textbf{Redundancy Type Definitions} for your analysis. If you identify a new redundancy type in Model A, you may include it in the output with a supplemental explanation.

\textbf{[Multi-Domain Translation: Redundancy Definitions]}
\begin{itemize}
\item \textbf{Over-segmentation}: Excessively decomposing simple sentences or obvious content into word-by-word or phrase-by-phrase fragments, causing unnecessary step inflation.
\item \textbf{Unnecessary linguistic explanation}: Over-explaining common sense or obvious linguistic points (grammar, part-of-speech, etymology, etc.) that do not influence the translation outcome.
\item \textbf{Semantic repetition}: Repeating the explanation of the same meaning or using different expressions to illustrate the same semantic point without adding value to the translation.
\item \textbf{Irrelevant information}: Introducing background knowledge, domain trivia, speculative content, or information unrelated to translation decisions (e.g., unnecessary context assumptions).
\item \textbf{Redundant alternative translations}: Providing multiple translation candidates with highly similar meanings and low contribution, or performing iterative optimization on trivial differences.
\item \textbf{Low-density long descriptions}: Using verbose sentences to express simple content or using a large volume of text to derive obvious reasoning steps.
\end{itemize}
\textbf{[Task Requirements]}
\begin{itemize}
\item Identify redundant snippets in Model A's CoT and label them with the corresponding redundancy type.
\item Explain why the snippet belongs to this category.
\item Determine if Model B's CoT eliminated this specific redundancy (Yes/No).
\item Output strictly as a structured JSON array in the following format:
\end{itemize}
\begin{verbatim}
[
  {
    "type": "Name of the redundancy type",
    "before_snippet": "Text snippet from Model A",
    "reason": "Reason for the judgment",
    "after_resolved": "Yes/No"
  },
  ...
]
\end{verbatim}

Now, please analyze the following two reasoning chains:
\newline
\textbf{Model A CoT:} \texttt{\{model\_a\_cot\}}
\newline
\textbf{Model B CoT:} \texttt{\{model\_b\_cot\}}

\end{tcolorbox}
\caption{The prompt used for the quantitative evaluation of reasoning efficiency.}
\label{fig:redundancy_prompt}
\end{figure*}

\begin{CJK}{UTF8}{gbsn}
\begin{table*}[t]
\centering
\footnotesize

\begin{tabular}{p{4.4cm}p{4.2cm}p{1.2cm}p{4.3cm}}
\toprule
\textbf{Source Sentence} & \textbf{Reference Translation} & \textbf{Difficulty} & \textbf{Reason} \\
\midrule
“我只是想要听听您的意见。”她用一副恭敬的口吻说道。 & “I just want to hear your opinion,” she said in a respectful tone. & \textbf{Easy} & A simple sentence with everyday vocabulary and direct grammar. No specialized knowledge required. \\
\addlinespace
At the same time the waves are fanning out, they are also separating by wavelength, a process known as dispersion. & 当这些波向外扩散时，它们也在按波长分开，这一过程叫频散。 & \textbf{Medium} & Contains a compound sentence and one scientific term (*dispersion*), requiring moderate structural and lexical control. \\
\addlinespace
Die folgende Tabelle zeigt die H\"aufigkeit und Schwere unerw\"unschter Wirkungen, die in >5\% von 168 Patienten mit Mesotheliom berichtet wurden... & The table below provides the frequency and severity of undesirable effects reported in $>$5\% of 168 mesothelioma patients... & \textbf{Hard} & Long and syntactically complex, includes multiple subordinate clauses, verb-final word order, and domain-specific medical terminology. Requires structural parsing and domain expertise. \\
\bottomrule
\end{tabular}
\caption{Detailed examples for translation difficulty evaluation.}
\label{tab:difficulty_example}
\end{table*}
\end{CJK}

\begin{figure*}[!htbp]
\centering
\includegraphics[width=\textwidth]{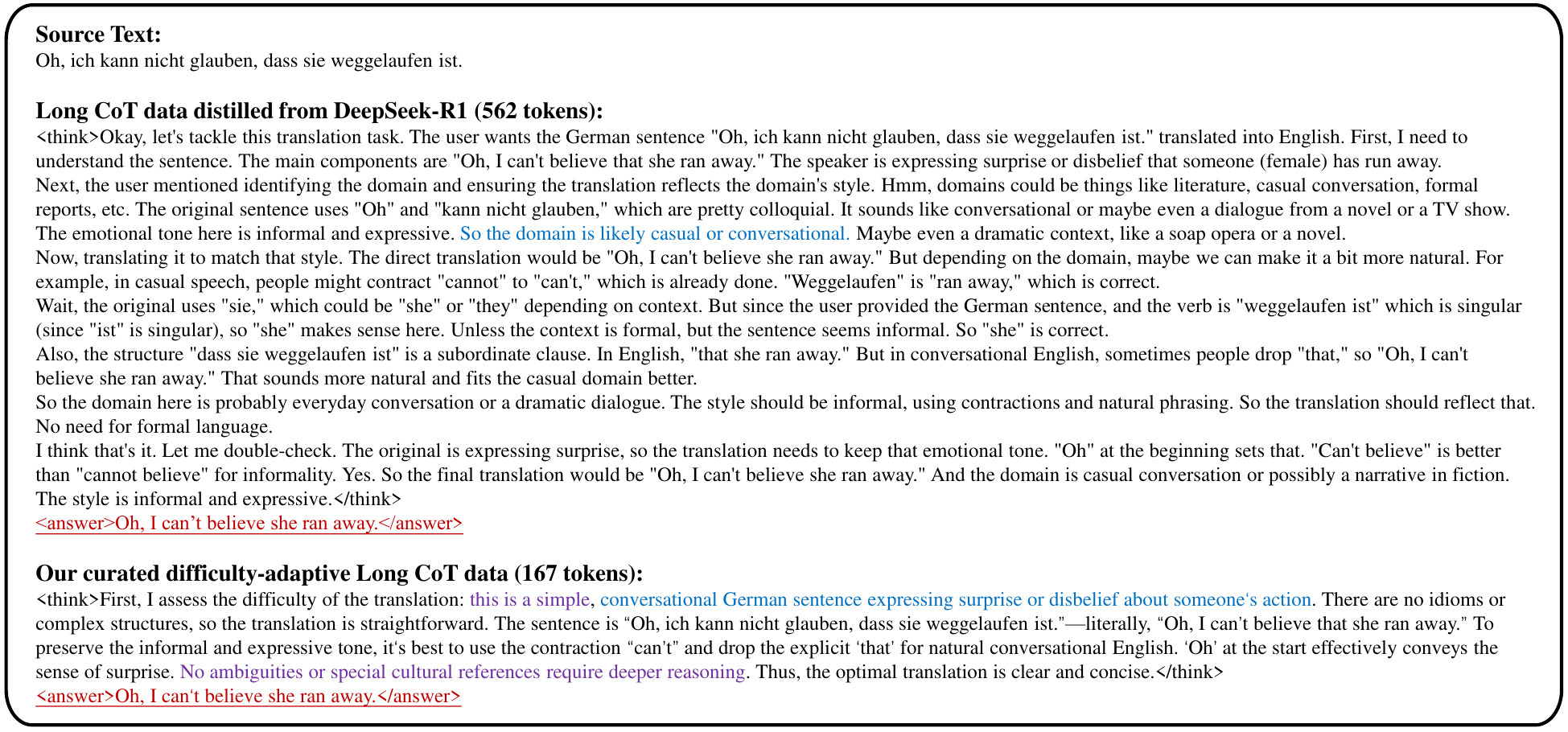}
\includegraphics[width=\textwidth]{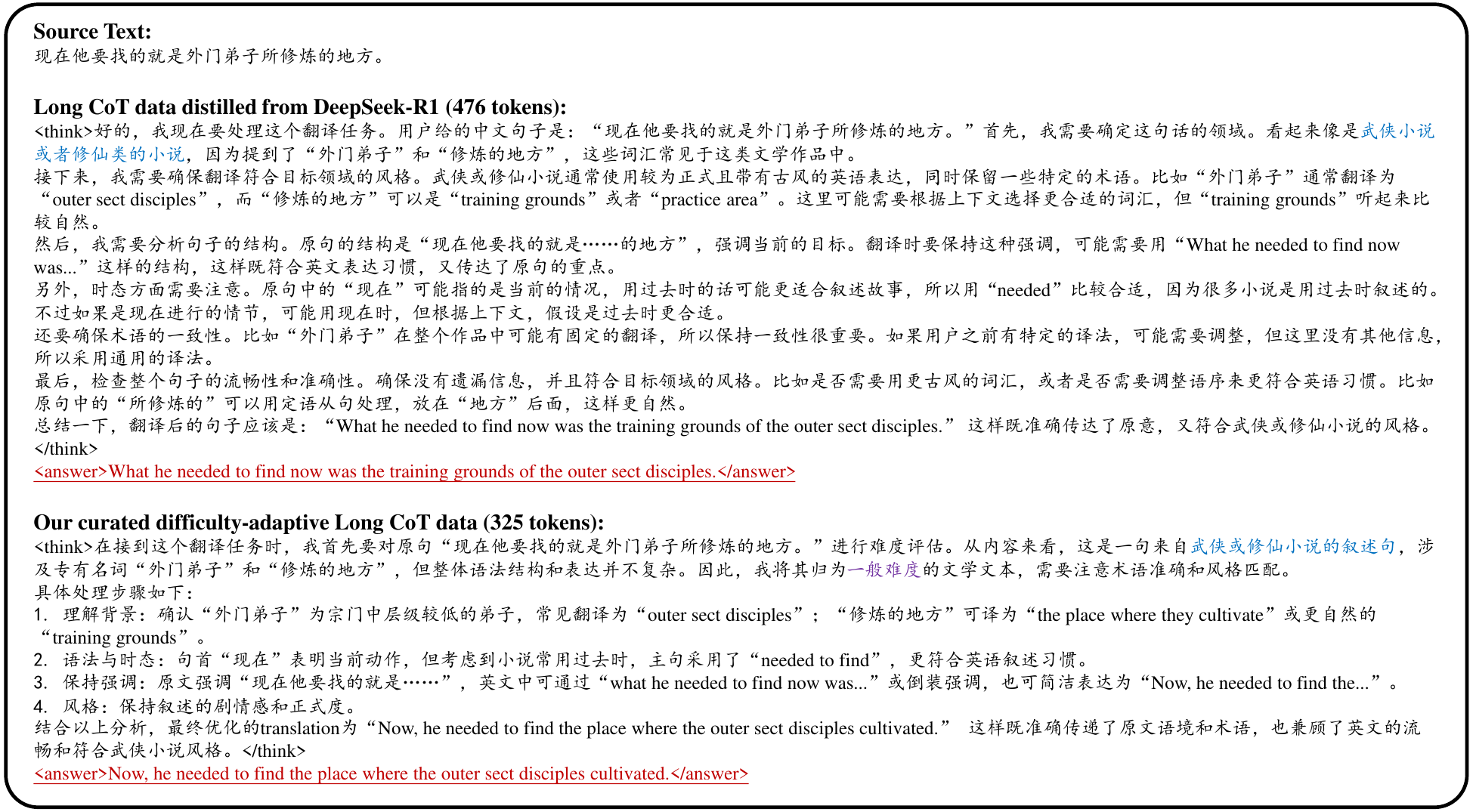}
\caption{Examples of our curated difficulty-adaptive Long CoT data (Easy and Medium).}
\label{fig:example_easy_mid}
\end{figure*}

\begin{figure*}[!htbp]
\centering
\includegraphics[width=\textwidth]{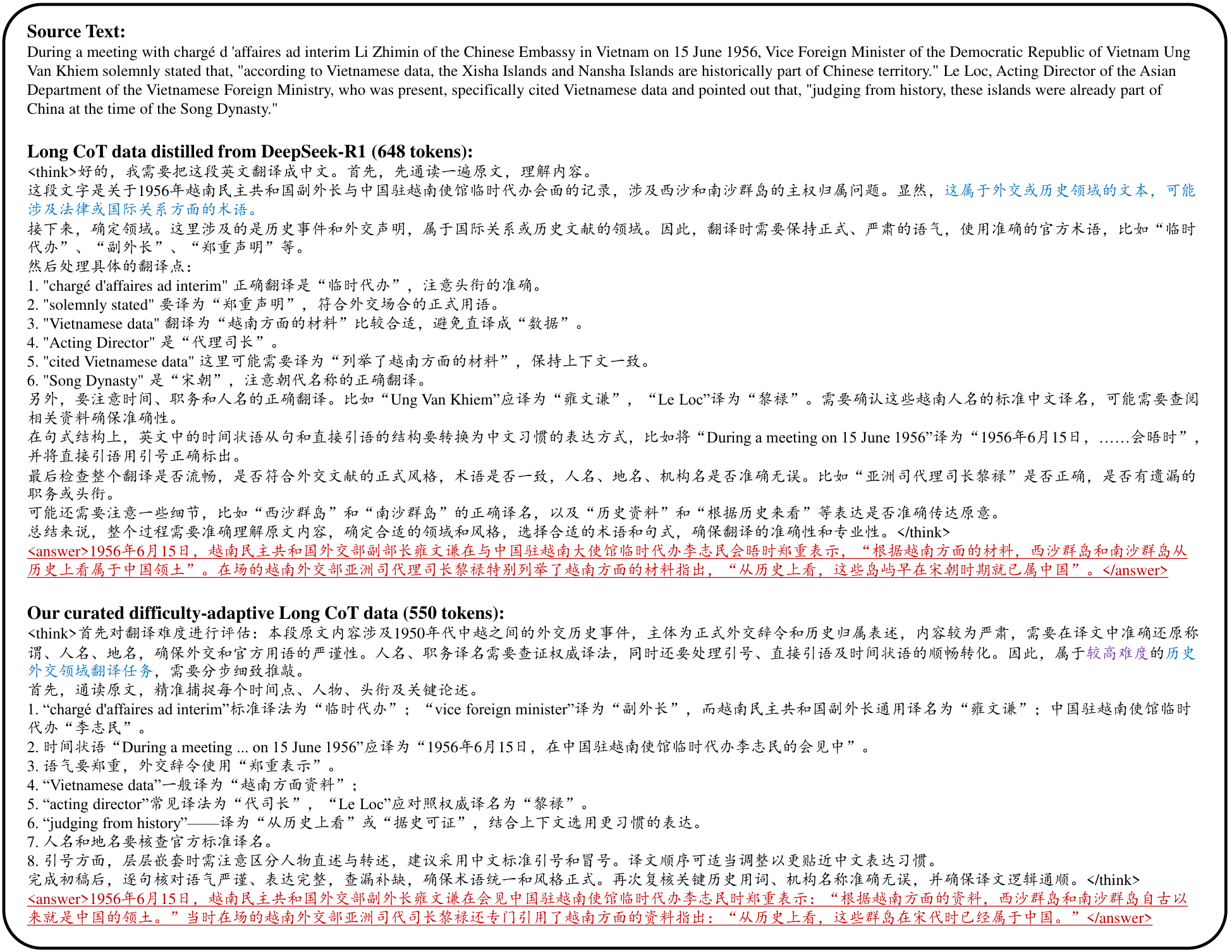}
\caption{Examples of our curated difficulty-adaptive Long CoT data (Hard).}
\label{fig:example_hard}
\end{figure*}

\begin{figure*}[!htbp]
\centering
\includegraphics[width=\textwidth]{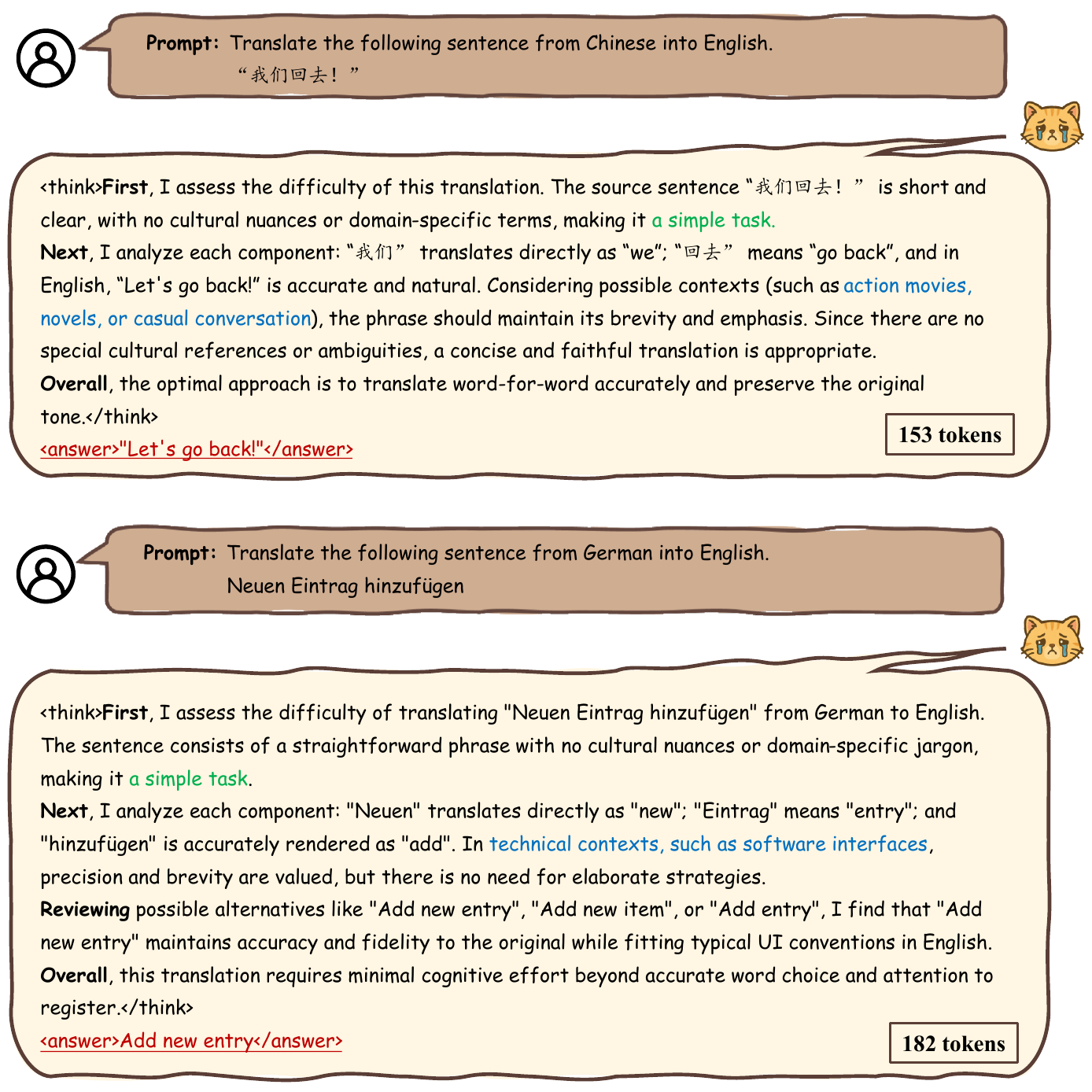}
\caption{Output examples of our \textbf{\texttt{TwT-7B}}.}
\label{fig:case_easy}
\end{figure*}

\begin{figure*}[!htbp]
\centering
\includegraphics[width=\textwidth]{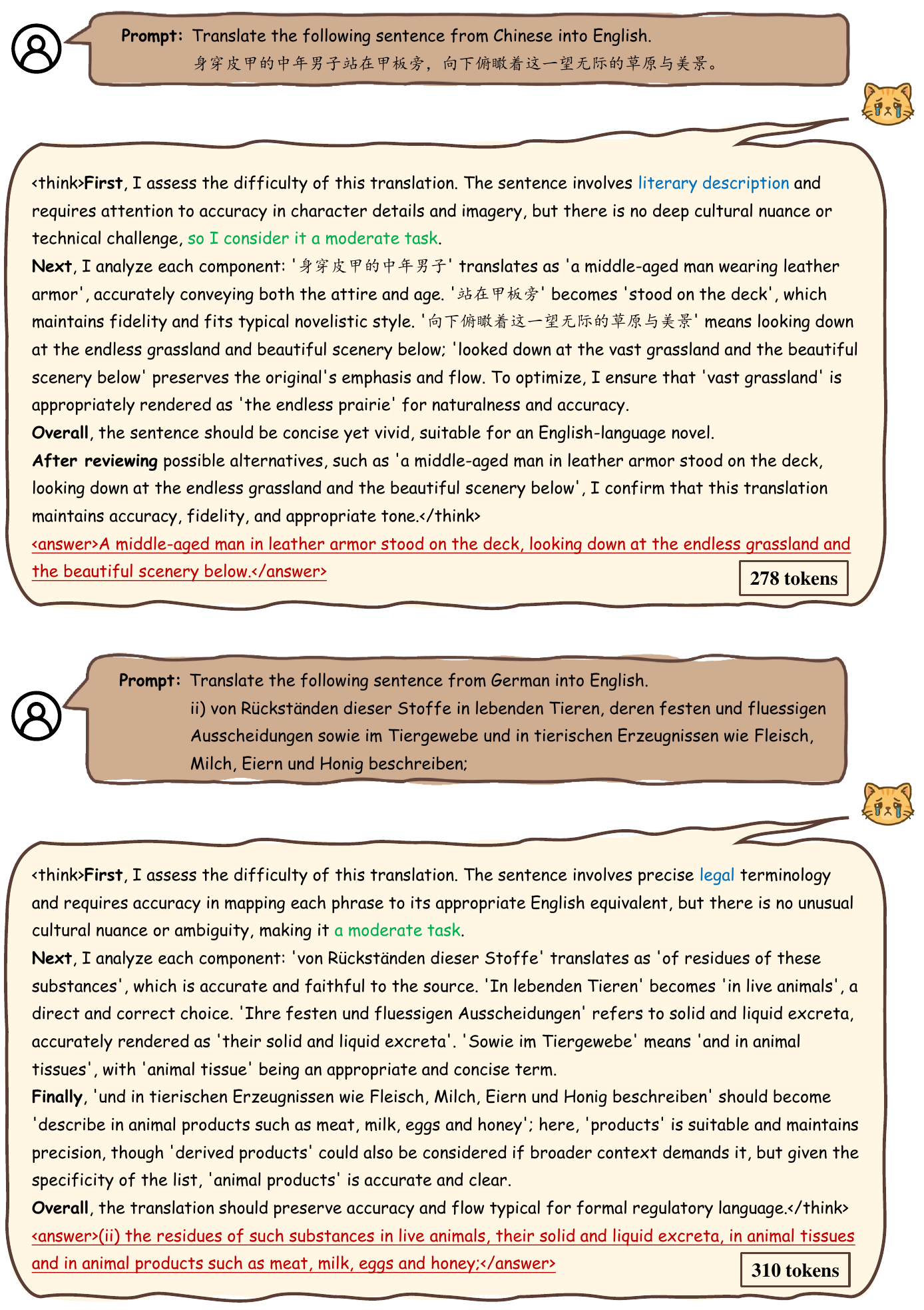}
\caption{Output examples of our \textbf{\texttt{TwT-7B}}.}
\label{fig:case_mid}
\end{figure*}

\section{Detailed Results by Metric}
\label{appendix:metric_breakdown}

For completeness, we provide the full breakdown of BLEU, COMET, and CometKiwi for the results reported in Table~\ref{tab:result_zh_en}, Table~\ref{tab:result_ood}, and Table~\ref{tab:result_lang}. The corresponding detailed results are shown in Table~\ref{tab:breakdown_in_domain}, Table~\ref{tab:breakdown_ood}, and Table~\ref{tab:breakdown_language}, respectively.

\begin{table*}[!ht]
\centering
\setlength{\tabcolsep}{9pt}
\resizebox{\textwidth}{!}{%
\begin{tabular}{lcccccccccccccccc}
\toprule
\multirow{2}{*}{\textbf{Method}} & \multicolumn{4}{c}{\textbf{Laws}} & \multicolumn{4}{c}{\textbf{News}} & \multicolumn{4}{c}{\textbf{Science}} & \multicolumn{4}{c}{\textbf{Subtitles}} \\
\cmidrule(lr){2-5} \cmidrule(lr){6-9} \cmidrule(lr){10-13} \cmidrule(lr){14-17}
 & \textbf{BLEU} & \textbf{COMET} & \textbf{KIWI} & \textbf{Tokens} & \textbf{BLEU} & \textbf{COMET} & \textbf{KIWI} & \textbf{Tokens} & \textbf{BLEU} & \textbf{COMET} & \textbf{KIWI} & \textbf{Tokens} & \textbf{BLEU} & \textbf{COMET} & \textbf{KIWI} & \textbf{Tokens} \\
\midrule

\rowcolor{mygray}
\multicolumn{17}{c}{\textbf{\textit{Large Language Models}}} \\
DeepSeek-V3 & \textbf{59.58} & {\ul 89.15} & 84.42 & - & {\ul 35.86} & \textbf{86.94} & 85.44 & - & \textbf{33.54} & \textbf{88.28} & 85.46 & - & 23.66 & 82.44 & 82.69 & - \\
Gemini-2.0-Flash & 56.43 & 88.90 & 84.39 & - & \textbf{36.39} & 86.37 & 85.14 & - & {\ul 33.47} & 87.69 & 85.24 & - & 24.64 & 81.91 & 82.33 & - \\
GPT-4o & 48.50 & 88.28 & 84.52 & - & 34.58 & 86.22 & 85.02 & - & 32.09 & 87.34 & 84.84 & - & 24.91 & 81.91 & 81.93 & - \\

\rowcolor{mygray}
\multicolumn{17}{c}{\textbf{\textit{Large Reasoning Models}}} \\
DeepSeek-R1 & {\ul 59.34} & {\ul 89.15} & 84.68 & 577 & 33.35 & 86.67 & 85.40 & 498 & 31.82 & {\ul 88.08} & {\ul 85.47} & 478 & 20.50 & 82.10 & {\ul 82.84} & 514 \\
Gemini-2.0-Flash-Thinking & 55.75 & 88.61 & 84.27 & 702 & 33.90 & 86.02 & 84.79 & 1149 & 31.98 & 87.47 & 85.06 & 1092 & 23.83 & 80.90 & 81.22 & 708 \\
OpenAI-o3-mini & 42.73 & 87.67 & 84.44 & 428 & 32.90 & 86.29 & 85.35 & 443 & 31.38 & 87.42 & 85.24 & 385 & 22.95 & 81.90 & 82.40 & 355 \\
OpenAI-o1 & 48.11 & 88.32 & {\ul 85.13} & 478 & 33.93 & 86.61 & {\ul 85.47} & 408 & 32.32 & 87.92 & \textbf{85.56} & 367 & 22.96 & 82.31 & 82.60 & 340 \\
GPT-5 & 54.06 & \textbf{89.34} & \textbf{85.43} & 784 & 35.05 & 86.65 & \textbf{85.52} & 740 & 32.21 & 88.06 & {\ul 85.47} & 606 & 23.23 & 82.32 & 82.68 & 519 \\
QwQ-32B & 43.77 & 87.35 & 84.39 & 667 & 33.24 & 85.77 & 84.63 & 584 & 32.05 & 87.11 & 84.42 & 563 & 22.58 & 81.54 & 81.29 & 584 \\

\rowcolor{mygray}
\multicolumn{17}{c}{\textbf{\textit{MT-Specialized Models}}} \\
SFT-Parallel-7B & 56.64 & 88.82 & 84.28 & - & 28.89 & 85.43 & 83.91 & - & 29.28 & 86.19 & 83.68 & - & {\ul 26.93} & 81.41 & 80.26 & - \\
ALMA-7B-R & 36.34 & 85.45 & 81.86 & - & 24.53 & 83.95 & 81.64 & - & 22.16 & 84.33 & 81.83 & - & 18.56 & 80.18 & 79.40 & - \\
ALMA-13B-R & 40.33 & 86.89 & 83.10 & - & 26.87 & 84.65 & 82.43 & - & 24.37 & 85.72 & 82.59 & - & 19.54 & 81.11 & 80.00 & - \\
TowerInstruct-7B-v0.2 & 50.53 & 88.25 & 82.96 & - & 30.72 & 84.61 & 82.45 & - & 27.76 & 85.60 & 82.99 & - & 22.45 & 80.75 & 80.01 & - \\
TowerInstruct-13B-v0.1 & 52.28 & 88.57 & 83.09 & - & 32.24 & 85.36 & 83.09 & - & 29.43 & 86.00 & 83.17 & - & 23.74 & 81.19 & 80.74 & - \\
CoT-FT-7B & 57.05 & 88.76 & 84.35 & 51 & 30.17 & 85.23 & 83.31 & 42 & 28.72 & 86.02 & 83.48 & 39 & \textbf{28.17} & 80.92 & 79.20 & 29 \\
MT-R1-Zero-7B & 35.49 & 86.78 & 84.51 & 72 & 31.69 & 86.01 & 84.54 & 64 & 29.45 & 86.87 & 84.61 & 61 & 22.48 & 81.73 & 81.70 & 55 \\
SSR-X-Zero-7B & 38.88 & 86.44 & 83.44 & 56 & 28.14 & 85.87 & 83.98 & 52 & 28.54 & 87.11 & 84.65 & 49 & 23.18 & 80.50 & 81.99 & 39 \\
mExTrans-7B & 38.77 & 87.11 & 84.44 & 597 & 25.05 & 86.07 & 85.34 & 553 & 25.74 & 87.34 & 85.31 & 546 & 14.31 & 81.40 & \textbf{82.85} & 476 \\

\rowcolor{mygray}
\multicolumn{17}{c}{\textbf{\textit{Our Models}}} \\
\textbf{\texttt{TwT-Qwen2.5-7B-Instruct}} & 52.56 & 88.76 & 84.73 & 310 & 33.63 & {\ul 86.75} & 84.88 & 311 & 32.41 & 87.53 & 84.77 & 294 & 24.74 & {\ul 82.63} & 81.73 & 247 \\
\textbf{\texttt{TwT-Qwen2.5-14B-Instruct}} & 56.03 & 89.07 & 84.65 & 320 & 34.57 & 86.65 & 84.63 & 285 & 32.75 & 87.54 & 84.66 & 272 & 24.44 & \textbf{82.73} & 81.74 & 241 \\

\bottomrule
\end{tabular}%
}

\vspace{0.6em}
\setlength{\tabcolsep}{5pt}
\resizebox{\textwidth}{!}{%
\begin{tabular}{lcccccccccccccccccccc}
\toprule
\multirow{2}{*}{\textbf{Method}} & \multicolumn{4}{c}{\textbf{Literary}} & \multicolumn{4}{c}{\textbf{IT}} & \multicolumn{4}{c}{\textbf{Koran}} & \multicolumn{4}{c}{\textbf{Medical}} & \multicolumn{4}{c}{\textbf{Average}} \\
\cmidrule(lr){2-5} \cmidrule(lr){6-9} \cmidrule(lr){10-13} \cmidrule(lr){14-17} \cmidrule(lr){18-21}
 & \textbf{BLEU} & \textbf{COMET} & \textbf{KIWI} & \textbf{Tokens} & \textbf{BLEU} & \textbf{COMET} & \textbf{KIWI} & \textbf{Tokens} & \textbf{BLEU} & \textbf{COMET} & \textbf{KIWI} & \textbf{Tokens} & \textbf{BLEU} & \textbf{COMET} & \textbf{KIWI} & \textbf{Tokens} & \textbf{BLEU} & \textbf{COMET} & \textbf{KIWI} & \textbf{Tokens} \\
\midrule

\rowcolor{mygray}
\multicolumn{21}{c}{\textbf{\textit{Large Language Models}}} \\
DeepSeek-V3 & 16.14 & 77.28 & 77.07 & - & 38.16 & 83.90 & 78.86 & - & 17.94 & 74.91 & 80.33 & - & 41.43 & 84.15 & 81.90 & - & 33.29 & 83.38 & 82.02 & - \\
Gemini-2.0-Flash & {\ul 18.37} & 77.20 & 76.67 & - & 37.93 & 83.20 & 78.49 & - & 19.70 & 74.97 & 79.71 & - & 44.39 & 84.49 & 81.76 & - & {\ul 33.91} & 83.09 & 81.72 & - \\
GPT-4o & 17.75 & 77.48 & 77.18 & - & 37.23 & 83.52 & 78.37 & - & 17.65 & 75.04 & 80.64 & - & 41.89 & 84.23 & 81.89 & - & 31.83 & 83.00 & 81.80 & - \\

\rowcolor{mygray}
\multicolumn{21}{c}{\textbf{\textit{Large Reasoning Models}}} \\
DeepSeek-R1 & 11.25 & 75.44 & 75.16 & 574 & 36.66 & 83.49 & 78.68 & 593 & 17.05 & 74.76 & 80.63 & 790 & 40.69 & 83.94 & 82.39 & 667 & 31.33 & 82.95 & 81.91 & 586 \\
Gemini-2.0-Flash-Thinking & 18.01 & 77.06 & 76.63 & 781 & 37.26 & 82.94 & 78.40 & 345 & 19.57 & 75.03 & 79.79 & 677 & 43.36 & 84.14 & 81.70 & 415 & 32.96 & 82.77 & 81.48 & 734 \\
OpenAI-o3-mini & 17.48 & 76.77 & 76.67 & 546 & 37.13 & 82.63 & 78.20 & 343 & 15.52 & 73.54 & 80.19 & 511 & 39.82 & 83.60 & 81.79 & 346 & 29.99 & 82.48 & 81.78 & 420 \\
OpenAI-o1 & 16.54 & 77.50 & \textbf{77.81} & 521 & 36.40 & 83.20 & {\ul 79.75} & 403 & 16.76 & 75.03 & \textbf{82.02} & 506 & 40.44 & 83.88 & \textbf{83.29} & 441 & 30.93 & 83.10 & \textbf{82.70} & 433 \\
GPT-5 & 15.01 & 76.72 & 77.08 & 859 & 36.92 & 83.58 & 79.22 & 492 & 18.83 & {\ul 75.49} & {\ul 81.80} & 751 & 43.00 & 84.29 & {\ul 83.13} & 531 & 32.29 & 83.31 & {\ul 82.54} & 660 \\
QwQ-32B & 12.99 & 75.98 & 76.31 & 863 & 21.93 & 82.95 & 79.43 & 583 & 12.36 & 73.59 & 81.47 & 963 & 33.95 & 83.27 & 82.82 & 735 & 26.61 & 82.20 & 81.85 & 693 \\

\rowcolor{mygray}
\multicolumn{21}{c}{\textbf{\textit{MT-Specialized Models}}} \\
SFT-Parallel-7B & 15.77 & 76.73 & 75.37 & - & {\ul 40.64} & {\ul 84.08} & 79.35 & - & 21.09 & 74.66 & 78.39 & - & 43.76 & 84.40 & 82.47 & - & 32.88 & 82.72 & 80.96 & - \\
ALMA-7B-R & 13.46 & 75.12 & 74.97 & - & 33.75 & 81.12 & 77.72 & - & 14.09 & 71.92 & 79.12 & - & 37.97 & 83.04 & 81.36 & - & 25.11 & 80.64 & 79.74 & - \\
ALMA-13B-R & 14.20 & 75.89 & 75.81 & - & 34.22 & 81.63 & 77.78 & - & 14.75 & 72.76 & 79.63 & - & 40.34 & 83.38 & 81.49 & - & 26.83 & 81.50 & 80.35 & - \\
TowerInstruct-7B-v0.2 & 15.54 & 75.49 & 74.93 & - & 38.20 & 83.47 & 78.67 & - & 10.99 & 69.02 & 70.13 & - & {\ul 46.81} & {\ul 84.58} & 80.80 & - & 30.38 & 81.47 & 79.12 & - \\
TowerInstruct-13B-v0.1 & 16.81 & 76.01 & 75.46 & - & 39.26 & 83.85 & 78.44 & - & 11.41 & 69.23 & 69.16 & - & \textbf{48.55} & \textbf{84.99} & 80.93 & - & 31.72 & 81.90 & 79.26 & - \\
CoT-FT-7B & 15.31 & 76.45 & 74.96 & 52 & 40.41 & 83.79 & 79.27 & 35 & 20.52 & 74.04 & 78.16 & 45 & 44.21 & 84.34 & 82.59 & 46 & 33.07 & 82.44 & 80.66 & 42 \\
MT-R1-Zero-7B & 13.75 & 76.93 & 76.72 & 69 & 34.51 & 82.71 & 79.59 & 56 & 13.05 & 72.84 & 80.66 & 71 & 27.28 & 83.31 & 82.69 & 71 & 25.96 & 82.15 & 81.88 & 65 \\
SSR-X-Zero-7B & 13.66 & 76.82 & 76.14 & 54 & 27.01 & 80.08 & 76.57 & 36 & 13.97 & 72.86 & 79.75 & 46 & 28.86 & 82.12 & 80.98 & 50 & 25.28 & 81.47 & 80.94 & 48 \\
mExTrans-7B & 10.43 & 76.16 & 76.62 & 610 & 25.54 & 78.76 & 77.19 & 452 & 11.80 & 73.88 & 80.56 & 604 & 25.40 & 81.66 & 81.93 & 565 & 22.13 & 81.55 & 81.78 & 551 \\

\rowcolor{mygray}
\multicolumn{21}{c}{\textbf{\textit{Our Models}}} \\
\textbf{\texttt{TwT-Qwen2.5-7B-Instruct}} & 17.93 & {\ul 78.35} & {\ul 77.30} & 281 & 40.12 & 84.04 & \textbf{80.05} & 222 & {\ul 21.13} & 75.45 & 80.19 & 269 & 43.74 & 84.32 & 82.81 & 262 & 33.28 & {\ul 83.48} & 82.06 & 274 \\
\textbf{\texttt{TwT-Qwen2.5-14B-Instruct}} & \textbf{19.10} & \textbf{78.40} & 76.93 & 354 & \textbf{41.26} & \textbf{84.17} & 79.68 & 234 & \textbf{22.20} & \textbf{75.71} & 80.13 & 336 & 44.45 & 84.42 & 82.89 & 287 & \textbf{34.35} & \textbf{83.59} & 81.91 & 291 \\

\bottomrule
\end{tabular}%
}
\caption{Detailed metric breakdown for Table~\ref{tab:result_zh_en}. We report in-domain translation results across eight domains, averaged over En$\rightarrow$Zh, Zh$\rightarrow$En, and De$\rightarrow$En.}
\label{tab:breakdown_in_domain}
\end{table*}

\begin{table*}[!ht]
\centering

\resizebox{\textwidth}{!}{%
\begin{tabular}{lcccccccccccc}
\toprule
\multirow{2}{*}{\textbf{Method}} & \multicolumn{4}{c}{\textbf{Conversation}} & \multicolumn{4}{c}{\textbf{Ecommerce}} & \multicolumn{4}{c}{\textbf{Social}} \\
\cmidrule(lr){2-5} \cmidrule(lr){6-9} \cmidrule(lr){10-13}
 & \textbf{BLEU} & \textbf{COMET} & \textbf{KIWI} & \textbf{Tokens} & \textbf{BLEU} & \textbf{COMET} & \textbf{KIWI} & \textbf{Tokens} & \textbf{BLEU} & \textbf{COMET} & \textbf{KIWI} & \textbf{Tokens} \\
\midrule

\rowcolor{gray!15}
\multicolumn{13}{c}{\textbf{\textit{Large Language Models}}} \\
DeepSeek-V3 & 36.76 & 87.03 & 81.48 & - & 32.29 & 85.53 & 81.19 & - & 32.26 & \textbf{84.59} & 81.44 & - \\
Gemini-2.0-Flash & \textbf{38.05} & 86.90 & 81.37 & - & {\ul 32.41} & 85.49 & 80.87 & - & \textbf{32.98} & 84.28 & 81.04 & - \\
GPT-4o & {\ul 38.02} & 86.79 & 81.43 & - & \textbf{32.88} & {\ul 85.57} & 81.04 & - & {\ul 32.75} & 84.47 & 81.08 & - \\

\rowcolor{gray!15}
\multicolumn{13}{c}{\textbf{\textit{Large Reasoning Models}}} \\
DeepSeek-R1 & 33.34 & 86.47 & 81.37 & 534 & 27.41 & 85.03 & 80.91 & 552 & 27.22 & 83.94 & 81.17 & 554 \\
Gemini-2.0-Flash-Thinking & 37.21 & 86.64 & 81.24 & 1204 & 31.50 & 85.24 & 80.74 & 822 & 31.79 & 83.89 & 80.83 & 1081 \\
OpenAI-o3-mini & 36.66 & 86.31 & 81.15 & 290 & 31.51 & 85.13 & 80.97 & 363 & 31.79 & 84.00 & 80.96 & 372 \\
OpenAI-o1 & 35.50 & {\ul 87.10} & {\ul 82.81} & 327 & 29.65 & \textbf{85.61} & \textbf{82.13} & 399 & 29.21 & {\ul 84.50} & \textbf{82.43} & 405 \\
GPT-5 & 35.25 & \textbf{87.14} & \textbf{82.82} & 448 & 29.09 & 85.37 & {\ul 82.02} & 609 & 29.00 & 84.10 & {\ul 82.31} & 652 \\

\rowcolor{gray!15}
\multicolumn{13}{c}{\textbf{\textit{MT-Specialized Models}}} \\
SFT-Parallel-7B & 32.20 & 84.11 & 80.62 & - & 26.94 & 82.92 & 80.12 & - & 25.88 & 81.61 & 80.12 & - \\
ALMA-7B-R & 29.49 & 84.67 & 79.96 & - & 25.13 & 82.98 & 79.50 & - & 25.76 & 82.51 & 79.77 & - \\
ALMA-13B-R & 31.86 & 85.48 & 80.76 & - & 26.24 & 83.75 & 80.12 & - & 27.00 & 83.06 & 80.47 & - \\
CoT-FT-7B & 31.58 & 84.09 & 80.63 & 31 & 26.71 & 82.95 & 80.11 & 45 & 24.98 & 81.37 & 79.83 & 42 \\
MT-R1-Zero-7B & 32.38 & 85.88 & 81.50 & 53 & 26.98 & 84.25 & 81.25 & 66 & 26.39 & 83.28 & 81.40 & 65 \\
SSR-X-Zero-7B & 30.65 & 85.32 & 81.13 & 37 & 25.64 & 83.94 & 80.79 & 50 & 25.47 & 83.19 & 81.26 & 49 \\
mExTrans-7B & 24.53 & 84.72 & 81.15 & 464 & 20.46 & 83.52 & 81.24 & 566 & 19.60 & 82.29 & 81.70 & 555 \\

\rowcolor{gray!15}
\multicolumn{13}{c}{\textbf{\textit{Our Models}}} \\
\textbf{\texttt{TwT-Qwen2.5-7B-Instruct}} & 35.10 & 86.37 & 81.69 & 231 & 30.92 & 84.85 & 81.35 & 273 & 30.79 & 84.08 & 81.58 & 269 \\
\textbf{\texttt{TwT-Qwen2.5-14B-Instruct}} & 35.35 & 86.39 & 81.58 & 240 & 31.04 & 85.11 & 81.21 & 309 & 31.04 & 83.91 & 81.46 & 298 \\

\bottomrule
\end{tabular}%
}

\vspace{0.6em}

\resizebox{\textwidth}{!}{%
\begin{tabular}{lcccccccccccc}
\toprule
\multirow{2}{*}{\textbf{Method}} & \multicolumn{4}{c}{\textbf{Culture}} & \multicolumn{4}{c}{\textbf{CommonSense}} & \multicolumn{4}{c}{\textbf{Average}} \\
\cmidrule(lr){2-5} \cmidrule(lr){6-9} \cmidrule(lr){10-13}
 & \textbf{BLEU} & \textbf{COMET} & \textbf{KIWI} & \textbf{Tokens} & \textbf{BLEU} & \textbf{COMET} & \textbf{KIWI} & \textbf{Tokens} & \textbf{BLEU} & \textbf{COMET} & \textbf{KIWI} & \textbf{Tokens} \\
\midrule

\rowcolor{gray!15}
\multicolumn{13}{c}{\textbf{\textit{Large Language Models}}} \\
DeepSeek-V3 & \textbf{40.27} & \textbf{85.46} & 83.21 & - & \textbf{32.55} & \textbf{85.36} & 79.98 & - & {\ul 34.83} & \textbf{85.59} & 81.46 & - \\
Gemini-2.0-Flash & {\ul 39.11} & 85.02 & 82.94 & - & 31.60 & 84.71 & 79.38 & - & {\ul 34.83} & 85.28 & 81.12 & - \\
GPT-4o & 38.65 & 85.24 & 83.14 & - & {\ul 32.54} & {\ul 85.26} & 79.87 & - & \textbf{34.97} & {\ul 85.47} & 81.31 & - \\

\rowcolor{gray!15}
\multicolumn{13}{c}{\textbf{\textit{Large Reasoning Models}}} \\
DeepSeek-R1 & 35.88 & {\ul 85.45} & 83.14 & 560 & 28.50 & 84.52 & 79.99 & 602 & 30.47 & 85.08 & 81.32 & 561 \\
Gemini-2.0-Flash-Thinking & 36.94 & 84.63 & 83.68 & 1220 & 32.15 & 84.50 & 79.68 & 2335 & 33.92 & 84.98 & 81.23 & 1332 \\
OpenAI-o3-mini & 34.05 & 84.36 & \textbf{84.25} & 596 & 28.23 & 84.28 & {\ul 80.85} & 436 & 32.45 & 84.82 & 81.64 & 411 \\
OpenAI-o1 & 34.57 & 85.06 & \textbf{84.25} & 542 & 28.43 & 84.97 & \textbf{80.95} & 392 & 31.47 & 85.45 & \textbf{82.52} & 413 \\
GPT-5 & 35.79 & 85.04 & {\ul 84.15} & 984 & 27.08 & 84.43 & 80.75 & 530 & 31.24 & 85.22 & {\ul 82.41} & 645 \\

\rowcolor{gray!15}
\multicolumn{13}{c}{\textbf{\textit{MT-Specialized Models}}} \\
SFT-Parallel-7B & 31.81 & 82.87 & 81.35 & - & 22.19 & 81.91 & 78.98 & - & 27.81 & 82.68 & 80.24 & - \\
ALMA-7B-R & 33.25 & 83.81 & 82.83 & - & 23.25 & 82.80 & 80.00 & - & 27.38 & 83.35 & 80.41 & - \\
ALMA-13B-R & 22.23 & 81.12 & 79.48 & - & 27.35 & 82.41 & 78.97 & - & 26.94 & 83.16 & 79.96 & - \\
CoT-FT-7B & 30.41 & 82.46 & 81.10 & 54 & 22.38 & 81.97 & 78.88 & 33 & 27.21 & 82.57 & 80.11 & 41 \\
MT-R1-Zero-7B & 31.86 & 83.99 & 82.85 & 79 & 24.50 & 82.94 & 79.53 & 51 & 28.42 & 84.07 & 81.31 & 63 \\
SSR-X-Zero-7B & 28.70 & 83.25 & 81.11 & 66 & 23.93 & 82.82 & 79.80 & 34 & 26.88 & 83.70 & 80.82 & 47 \\
mExTrans-7B & 27.12 & 84.30 & 83.92 & 631 & 18.61 & 81.92 & 79.23 & 470 & 22.06 & 83.35 & 81.45 & 537 \\

\rowcolor{gray!15}
\multicolumn{13}{c}{\textbf{\textit{Our Models}}} \\
\textbf{\texttt{TwT-Qwen2.5-7B-Instruct}} & 35.24 & 84.59 & 83.64 & 352 & 29.08 & 84.30 & 80.07 & 219 & 32.23 & 84.84 & 81.67 & 269 \\
\textbf{\texttt{TwT-Qwen2.5-14B-Instruct}} & 37.70 & 84.71 & 83.23 & 333 & 29.81 & 84.36 & 79.98 & 259 & 32.99 & 84.90 & 81.49 & 288 \\

\bottomrule
\end{tabular}%
}
\caption{Detailed metric breakdown for Table~\ref{tab:result_ood}. We report OOD translation results across five domains, averaged over En$\rightarrow$Zh, Zh$\rightarrow$En, and De$\rightarrow$En.}
\label{tab:breakdown_ood}
\end{table*}
\begin{table*}[t]
\centering
\resizebox{\textwidth}{!}{%
\begin{tabular}{lcccccccccccc}
\toprule
\multirow{2}{*}{\textbf{Method}} & \multicolumn{4}{c}{\textbf{En$\rightarrow$Zh}} & \multicolumn{4}{c}{\textbf{Zh$\rightarrow$En}} & \multicolumn{4}{c}{\textbf{De$\rightarrow$En}} \\
\cmidrule(lr){2-5} \cmidrule(lr){6-9} \cmidrule(lr){10-13}
 & \textbf{BLEU} & \textbf{COMET} & \textbf{KIWI} & \textbf{Tokens} & \textbf{BLEU} & \textbf{COMET} & \textbf{KIWI} & \textbf{Tokens} & \textbf{BLEU} & \textbf{COMET} & \textbf{KIWI} & \textbf{Tokens} \\
\midrule

\rowcolor{gray!15}
\multicolumn{13}{c}{\textbf{\textit{Large Language Models}}} \\
Qwen2.5-7B-Instruct & 35.05 & 85.44 & 83.11 & - & 12.63 & 69.88 & 70.22 & - & 28.82 & 79.36 & 80.53 & - \\
Gemma-2-9B-IT & 32.66 & 84.49 & 82.36 & - & 13.57 & 72.41 & 71.94 & - & 25.84 & 77.06 & 79.01 & - \\

\rowcolor{gray!15}
\multicolumn{13}{c}{\textbf{\textit{MT-Specialized Models}}} \\
ALMA-7B-R & 26.69 & 84.13 & 81.43 & - & 13.46 & 75.12 & 74.97 & - & 28.87 & 79.66 & 80.13 & - \\
Tower-Plus-9B & 37.91 & 86.76 & {\ul 84.88} & - & 16.56 & 77.88 & \textbf{77.78} & - & {\ul 36.56} & \textbf{82.21} & \textbf{82.36} & - \\
SFT-Parallel-7B & \textbf{38.58} & 86.34 & 83.39 & - & 15.77 & 76.73 & 75.37 & - & 35.72 & 81.63 & 81.11 & - \\
MT-R1-Zero-7B & 32.00 & 86.23 & 84.34 & 62 & 13.75 & 76.93 & 76.72 & 69 & 25.56 & 80.39 & 81.73 & 66 \\
SSR-X-Zero-7B & 34.20 & {\ul 86.86} & 84.58 & 50 & 14.83 & 77.45 & 77.27 & 54 & 26.65 & 80.39 & 81.55 & 45 \\
mExTrans-7B & 28.63 & 86.44 & \textbf{85.02} & 537 & 10.43 & 76.16 & 76.62 & 610 & 22.06 & 79.27 & 80.88 & 544 \\

\rowcolor{gray!15}
\multicolumn{13}{c}{\textbf{\textit{Our Models}}} \\
\textbf{\texttt{TwT-Qwen2.5-7B-Instruct}} & {\ul 38.28} & \textbf{87.27} & 84.43 & 298 & {\ul 17.93} & \textbf{78.35} & {\ul 77.30} & 281 & 35.53 & 81.93 & 81.90 & 256 \\
\textbf{\texttt{TwT-Gemma-2-9B-IT}} & 36.34 & 86.65 & 84.23 & 227 & \textbf{19.17} & {\ul 78.11} & 77.07 & 249 & \textbf{36.73} & {\ul 82.15} & {\ul 81.97} & 218 \\

\bottomrule
\end{tabular}%
}

\vspace{0.6em}

\resizebox{\textwidth}{!}{%
\begin{tabular}{lcccccccccccc}
\toprule
\multirow{2}{*}{\textbf{Method}} & \multicolumn{4}{c}{\textbf{En$\rightarrow$X}} & \multicolumn{4}{c}{\textbf{X$\rightarrow$En}} & \multicolumn{4}{c}{\textbf{Average}} \\
\cmidrule(lr){2-5} \cmidrule(lr){6-9} \cmidrule(lr){10-13}
 & \textbf{BLEU} & \textbf{COMET} & \textbf{KIWI} & \textbf{Tokens} & \textbf{BLEU} & \textbf{COMET} & \textbf{KIWI} & \textbf{Tokens} & \textbf{BLEU} & \textbf{COMET} & \textbf{KIWI} & \textbf{Tokens} \\
\midrule

\rowcolor{gray!15}
\multicolumn{13}{c}{\textbf{\textit{Large Language Models}}} \\
Qwen2.5-7B-Instruct & 5.97 & 55.56 & 51.30 & - & 20.45 & 76.22 & 74.30 & - & 20.58 & 73.29 & 71.89 & - \\
Gemma-2-9B-IT & \textbf{14.05} & \textbf{75.63} & {\ul 73.16} & - & {\ul 31.55} & {\ul 84.71} & {\ul 82.88} & - & 23.53 & {\ul 78.86} & 77.87 & - \\

\rowcolor{gray!15}
\multicolumn{13}{c}{\textbf{\textit{MT-Specialized Models}}} \\
ALMA-7B-R & 2.78 & 58.13 & 73.13 & - & 12.01 & 63.94 & 58.17 & - & 16.76 & 72.20 & 73.57 & - \\
Tower-Plus-9B & 6.07 & 63.86 & 69.46 & - & 27.99 & 81.38 & 79.68 & - & {\ul 25.02} & 78.42 & {\ul 78.83} & - \\
SFT-Parallel-7B & 1.19 & 52.16 & 43.97 & - & 18.47 & 75.58 & 73.05 & - & 21.95 & 74.49 & 71.38 & - \\
MT-R1-Zero-7B & 5.91 & 58.76 & 58.27 & 358 & 21.91 & 77.48 & 75.61 & 74 & 19.82 & 75.96 & 75.34 & 126 \\
SSR-X-Zero-7B & 6.02 & 58.94 & 56.34 & 306 & 20.73 & 78.05 & 75.78 & 48 & 20.49 & 76.34 & 75.10 & 101 \\
mExTrans-7B & 5.09 & 61.33 & 61.74 & 1047 & 15.95 & 76.80 & 75.21 & 731 & 16.43 & 76.00 & 75.89 & 694 \\

\rowcolor{gray!15}
\multicolumn{13}{c}{\textbf{\textit{Our Models}}} \\
\textbf{\texttt{TwT-Qwen2.5-7B-Instruct}} & 5.99 & 59.57 & 57.83 & 483 & 22.94 & 77.73 & 75.84 & 328 & 24.13 & 76.97 & 75.46 & 329 \\
\textbf{\texttt{TwT-Gemma-2-9B-IT}} & {\ul 11.12} & {\ul 72.02} & \textbf{80.82} & 280 & \textbf{32.79} & \textbf{85.02} & \textbf{83.30} & 257 & \textbf{27.23} & \textbf{80.79} & \textbf{81.48} & 246 \\

\bottomrule
\end{tabular}%
}
\caption{Detailed metric breakdown for Table~\ref{tab:result_lang}. We report results on seen and unseen language directions. En, Zh, and De are \emph{seen} languages, while X denotes \emph{unseen} languages; En$\rightarrow$X and X$\rightarrow$En report averages over English$\leftrightarrow$unseen-language directions.}
\label{tab:breakdown_language}
\end{table*}

\end{document}